\documentclass[10pt,twocolumn,letterpaper]{article}

\usepackage[pagenumbers]{cvpr} 

\usepackage{lineno}
\definecolor{cvprblue}{rgb}{0.21,0.49,0.74}
\usepackage[pagebackref,breaklinks,colorlinks,allcolors=cvprblue]{hyperref}
\usepackage{xcolor}
\usepackage{graphicx}
\usepackage[table]{xcolor}
\usepackage{pgf}
\usepackage{float}
\usepackage{fontawesome5}

\newcommand{\rgscalemax}{1.7} 

\newcommand{\rgscore}[2]{%
  \begingroup
  \pgfmathsetmacro{\s}{#1}
  \pgfmathsetmacro{\amax}{\rgscalemax}
  \pgfmathparse{min(100,max(0,round(100*abs(\s)/\amax)))}%
  \let\pct\pgfmathresult
  \pgfmathparse{\s<0 ? 1 : (\s>0 ? 0 : 2)}\let\sgn\pgfmathresult
  \def\bg{white}%
  \ifnum\sgn=1\edef\bg{red!\pct}\fi
  \colorbox{\bg}{\textcolor{black}{#2}}%
  \endgroup
}

\newcommand{\bluehlscalemax}{130}
\newcommand{\bluehl}[2]{%
  \begingroup
  \pgfmathsetmacro{\amax}{\bluehlscalemax}%
  \pgfmathparse{100*abs(#1)/\amax}%
  \pgfmathtruncatemacro{\pct}{min(100,max(0,round(\pgfmathresult)))}%
  \ifnum\pct=0\relax
    #2%
  \else
    \colorbox{orange!\pct}{#2}%
  \fi
  \endgroup
}

\usepackage[ruled,vlined,linesnumbered]{algorithm2e}
\usepackage{amsmath}

\usepackage{booktabs,threeparttable,multirow}

\def\paperID{*****} 
\def\confName{CVPR}
\def\confYear{2026}

\usepackage{booktabs}

\newcommand{\confusedtourist}{\textsc{ConfusedTourist }}

\title{Vision Language Models are Confused Tourists  \raisebox{-0.2em}{\includegraphics[height=1.2em]{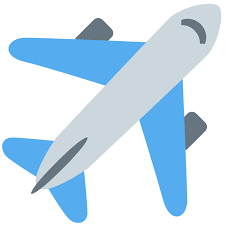}}}

\author{Patrick Amadeus Irawan$^{1*}$, Ikhlasul Akmal Hanif$^{1*}$, Muhammad Dehan Al Kautsar$^1$,\\
Genta Indra Winata$^{2\dagger}$, Fajri Koto$^{1\dagger}$, Alham Fikri Aji$^{1\dagger}$\\
$^1$MBZUAI$\quad$$^2$Capital One\\
{\footnotesize{$^\ast$Main Authors, $^\dagger$Senior Authors}}\\
{\tt\small ({patrick.irawan, ikhlasul.hanif})@mbzuai.ac.ae}\\[0.5em]
\href{https://github.com/patrickamadeus/vlms-are-confused-tourists}{\color{black}\faGithub\ \textbf{Code}} \quad
\href{https://huggingface.co/datasets/patrickamadeus/vlms-are-confused-tourists}{\color{black}\raisebox{-0.4em}{\includegraphics[height=1.6em]{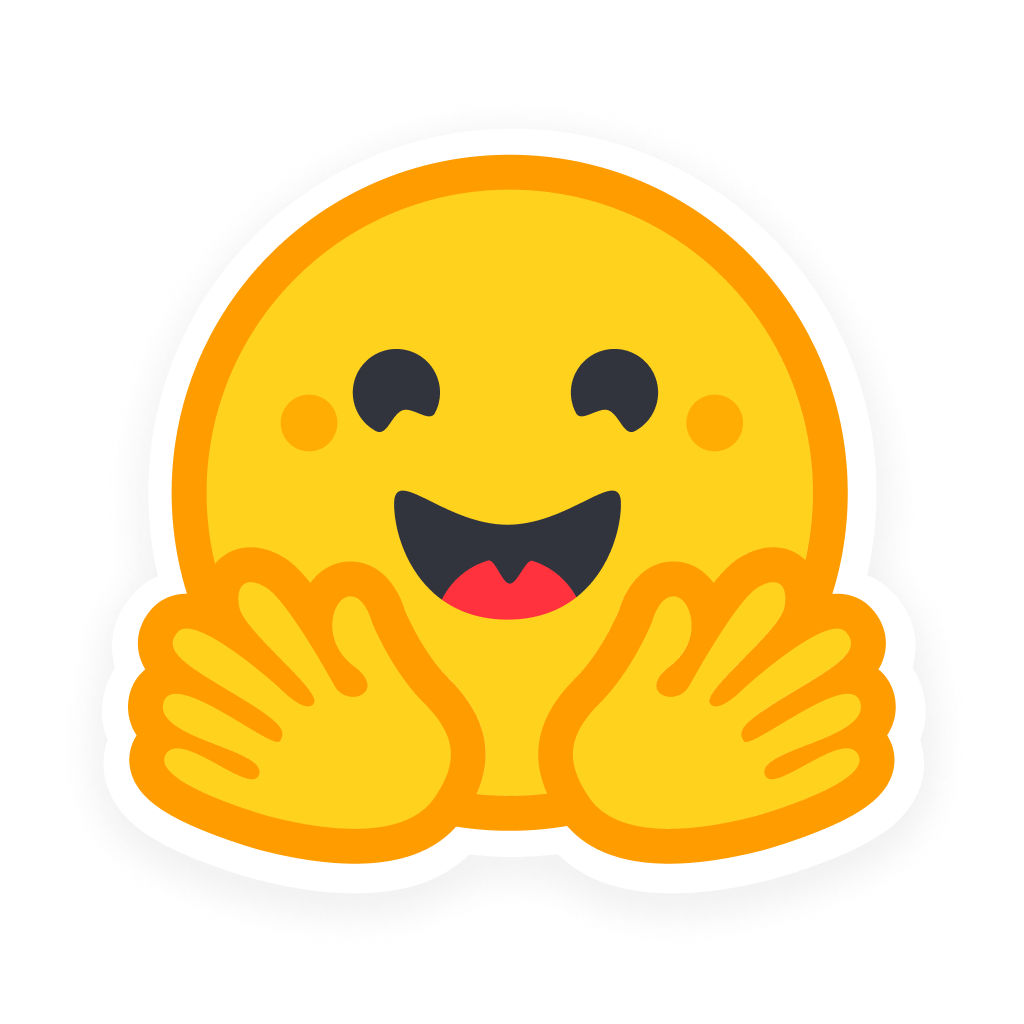}}\ \textbf{Dataset}} \quad
}

\usepackage{capt-of,etoolbox}
\makeatletter
\newcommand\mytopfigure{%
  \begin{center}%
    \includegraphics[width=0.9\linewidth]{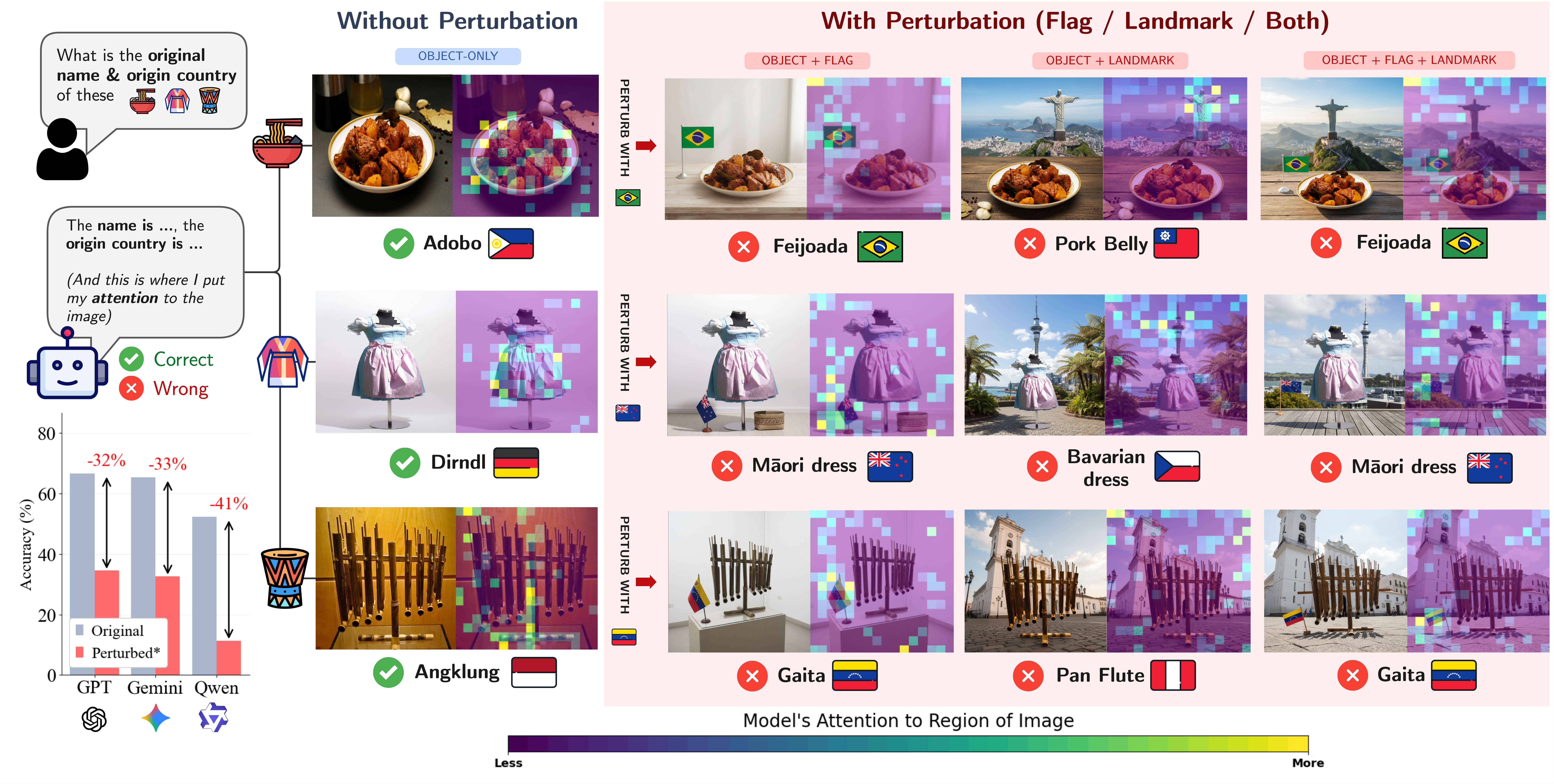}%
    \captionof{figure}{\confusedtourist evaluates the robustness of current state-of-the-art VLMs on grounding single cultural concept within images that contain distracting cue(s). We demonstrate how geographical-induced perturbation causes massive accuracy drop (flag-perturbation$^*$) consistently across all cases. Through further interpretability analysis, we also discover that the model's distractive attention shift  toward the adversarial cue(s) (e.g., Row 1 example where the Brazil flag was attended significantly more than the Adobo cuisine) can directly explain the decline.}
    \label{fig:placeholder}%
  \end{center}%
}
\apptocmd\@maketitle{{\mytopfigure\par}}{}{}
\makeatother

\begin{document}
\maketitle


\begin{abstract}

Although the cultural dimension has been one of the key aspects in evaluating Vision-Language Models (VLMs), their ability to remain stable across diverse cultural inputs remains largely untested, despite being crucial to support diversity and multicultural societies. Existing evaluations often rely on benchmarks featuring only a singular cultural concept per image, overlooking scenarios where multiple, potentially unrelated cultural cues coexist. To address this gap, we introduce \confusedtourist, a novel cultural adversarial robustness suite designed to assess VLMs’ stability against perturbed geographical cues. Our experiments reveal a critical vulnerability, where accuracy drops heavily under simple image-stacking perturbations and even worsens with its image-generation-based variant. Interpretability analyses further show that these failures stem from systematic attention shifts toward distracting cues, diverting the model from its intended focus. These findings highlight a critical challenge: visual cultural concept mixing can substantially impair even state-of-the-art VLMs, underscoring the urgent need for more culturally robust multimodal understanding.

\end{abstract}
\section{Introduction}
\label{sec:intro}


Recent advances in multimodality have enabled Vision-Language Models (VLMs) to become more proficient across a range of tasks, including ones that require domain-specific knowledge. Multicultural domain fits naturally into such description, as it requires models to have a culturally specific understanding to be able to capture relevant insight from rich visual inputs. Prior benchmarks have set the groundwork for such evaluation, including in the general-purpose scope \cite{romero2024cvqa, vayani2024almbench, culturevqa2024, satar2025seeingculturebenchmarkvisual} and in a more fine-grained category setting (e.g. cuisine-only \cite{winata-etal-2025-worldcuisines, li2024foodieqamultimodaldatasetfinegrained} or paintings-only \cite{zhang2025cultiverse}). These systematic large-scale benchmarks were built to assess whether VLMs can probe and reason about multicultural knowledge beyond basic visual perception.

Although recent numbers\footnote{We evaluated recent frontier VLMs to verify this observation; detailed results are provided in Appendix~\ref{appendix:eval-prior}.} indicate collective improvement of recent VLMs' multicultural comprehension ability, aforementioned benchmarks only assess this understanding using unambiguous cultural images. These images typically contain either only a single concept (e.g., a person wearing a Japanese kimono) or multiple but inherently related cues (e.g., Indonesian Gamelan musicians playing drums and gongs next to temple dancers), thereby allowing grounding inference to be assisted by the existence of related cues that may not refer to the intended concept. This limitation makes it difficult to disentangle whether the VLM is identifying an object based on its intrinsic visual features or if it is overly relying on other contextual cues. Hence, the ideal setting for stress-testing this robustness is by involving scenes with simultaneously contrasting cultural cues to challenge the model's ability to refer to relevant object.





Prior works have attempted similar perturbation ideas in the multicultural domain. For instance, \citet{ye-etal-2025-claim} conducted a text-modality perturbation check on cross-modal attention divergence in a cross-lingual setting, indicating that image understanding levels highly differ between languages. On the other modality side, \citet{kimchi2025} attempted to perturb the image by altering the ethnicity of persons to point out potential bias of VLMs. However, these works suffer from one or more of the following limitations: \textbf{(1)} primarily involve adversity just in text, \textbf{(2)} involve concepts that are inherently subjective or culturally biased, making explicit and objective analysis difficult and potentially leading to uneven treatment of different concepts (for example: racial or cultural groups in \citet{kimchi2025}), and \textbf{(3)} did not include a more comprehensive analysis over the model's failure behavior.



In this work, we address these critical gaps by introducing a novel evaluation suite designed to probe whether state-of-the-art VLMs are able to identify a specific cultural item amidst the existence of other perturbing cues. To achieve this, we construct a suite of test images where a target cultural item is deliberately co-presented with a conflicting geographical symbol or object (such as a flag or landmark). We include rich multicultural nuances across various cultural domains, concepts, perturbation contexts, image creation methods, together with empirical and behavioral analyses of notable failure cases. Our contributions are summarized as follows:

\begin{enumerate}
    \item We present $\confusedtourist$, a VL robustness evaluation suite comprising of 5k+ geographical-cue perturbed images which includes 243 unique culture items from 57 countries. We curate the altered images by applying 3 perturbation settings (flag-only, landmark-only, or both) on multiple difficulties.
    

    \item We benchmark 14 SOTA VLMs on our suite, where we observe a consistent and critical performance degradation in all cases, even showing failure against the simple perturbing method ($\ref{par:image_stacking}$).
   
    \item We provide further interpretability analyses of open-source VLMs, highlighting their attention-shift behavior, biases, and reasoning thought to explain the nature of this performance drop.
\end{enumerate}

\section{Related Work}

\paragraph{Multicultural Understanding in Vision Language.}
\label{sec:rel_multi_cultural}
Prior work has shown that VLMs exhibit cultural blind spots due to training data heavily favoring Western contexts. Recent multicultural VLMs benchmarks ~\cite{liu2025culturevlm, romero2024cvqa, winata-etal-2025-worldcuisines, Vayani_2025_CVPR} consistently demonstrate that VLMs struggle to accurately recognize artifacts, foods, and traditions from non-Western or underrepresented regions.
Studies such as~\citet{ye-etal-2025-claim} examine how text-modality perturbations affect cross-modal attention in cross-lingual settings, while ~\citet{kimchi2025} reveal that VLM predictions can be improperly influenced by irrelevant cues, such as the perceived ethnicity of a person in the image. Other works further confirm these challenges in multicultural grounding~\cite{culturevqa2024, satar2025seeingculturebenchmarkvisual}.
Our research extends this line of inquiry by examining how a VLM’s cultural understanding behaves under conflicting visual cues, addressing a critical and previously unexplored aspect of multicultural robustness.


\paragraph{VLM Robustness in Concept-Mixed Setting.}
\label{sec:rel_robustness}

VLM robustness research typically addresses two failure categories: failures against simple noise and failures against semantic complexity. Traditional studies focus on out-of-distribution corruption, such as added noise or common visual distortions \cite{hendrycks2019benchmarkingneuralnetworkrobustness, hendrycks2021many}, which test models' basic perceptual stability \cite{bhojanapalli2021understanding, vo2025vision, aravindan2025vlms,rahmanzadehgervi2025visionlanguagemodelsblind}. However, VLMs suffer from distinct vulnerabilities related to higher-level meaning and concept mixing. Failures have been observed in geometric reasoning \cite{rahmanzadehgervi2024vision}, compositional generalization \cite{pearson2025evaluatingcompositionalgeneralisationvlms, aravindan2025vlms, ismayilzada-etal-2025-evaluating}, and bias driven by strong visual or linguistic priors \cite{vo2025vision}. Furthermore, models frequently misattribute facts and hallucinate plausible but incorrect information, demonstrating weak factual grounding \cite{rahmanzadehgervi2024vision}. While prior work exists in multicultural settings \cite{ye-etal-2025-claim, kimchi2025}, these efforts fail to convey a detailed robustness study, especially one involving contrasting and non-subjective cultural concepts. Our study bridges this gap by treating cross-cultural visual mixing as a form of semantically guided perturbation, analyzing how VLMs fail to maintain factual focus when multiple, conflicting cultural concepts coexist.

      

\begin{table}[t]
  \centering
  \small 
  \setlength{\tabcolsep}{3pt} 
  \begin{tabular}{l c c c c c c}
    \toprule
    & & \multicolumn{2}{c}{\textbf{Desc.}} & \multicolumn{2}{c}{\textbf{Geo.}} & \\
    \cmidrule(lr){3-4}\cmidrule(lr){5-6}
    \textbf{Category} & \textbf{Ori.} &
    \textbf{Easy} & \textbf{Hard} &
    \textbf{Easy} & \textbf{Hard} &
    \textbf{Total Pairs} \\
    \midrule
    Cuisine (181) & 181
      & 894 & 1038 & 912 & 912 & 3756 \\
    Attire (38) & 38
      & 228 & 228 & 216 & 216 & 888 \\
    Music (24) & 24
      & 144 & 144 & 138 & 138 & 564 \\
    \midrule
    \textbf{Grand Total} & 243 & 1266 & 1410 & 1266 & 1266 & 5451 \\
    \bottomrule
  \end{tabular}
  \caption{Dataset statistics for $\confusedtourist$. With description- or geographical-based pairs, each image is perturbed at 2 difficulty levels using image stacking or generative perturbation. A more detailed stat of the suite can be seen in the Appendix.}
  \label{tab:culture_perturb_stats_gen}
\end{table}
\section{\confusedtourist Evaluation Suite}
\label{sec:methodology}

\begin{figure*}
    \centering
    \includegraphics[width=\linewidth]{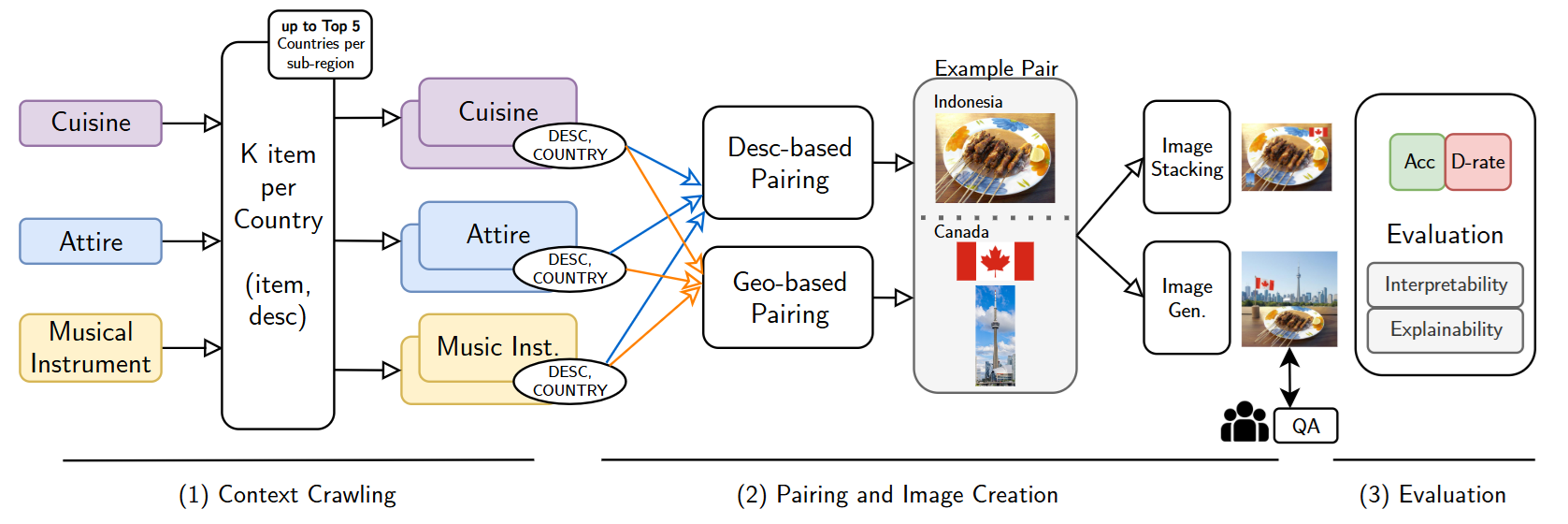}
    \caption{Our \confusedtourist construction pipeline. The pipeline consists of 3 stages: \textbf{(1) Context Crawling} to obtain balanced, culturally diverse item data and descriptions; \textbf{(2) Pair \& image creation} where we generate hard and easy cultural pairings and produce various perturbation-infused visual cases; and \textbf{(3) Evaluation}, where we assess VLMs' concept grounding ability using objective metrics and interpretability analysis.}
    \label{fig:data-method}
\end{figure*}

Our suite is comprised of 5,451 unique images, featuring 243 unique cultural items sourced from 57 countries across 11 sub-regions. The suite cover 3 categories, including cuisine, traditional attire, and musical instruments. We curate this suite using a 3-staged pipeline involving context crawling, pair creation, and image generation, with details depicted in Figure \ref{fig:data-method}. Table \ref{tab:culture_perturb_stats_gen} presents the overall statistics of our data.

\subsection{Context Crawling}
\label{sec:context_crawl}

\paragraph{Cultural Domain.} We select cuisine, traditional attire, and musical instruments as our categories because they meet two criteria: they are well-constrained and their items are typically represented by a single, clear, object-based cue. This approach allows us to reduce the ambiguity found in multi-concept categories (like festivals and games) or overly broad categories where items take many forms (like artifacts or gifts).



\paragraph{Geographical Context.}
Our country selection covers 57 countries sampled from 11 sub-regions, with each sub-region containing a maximum of 7 countries. We design this sampling strategy to balance sub-regional diversity, rather than focusing on larger divisions like continents. For each country's visual perturbation image grounding resources, we obtained flag and landmark images from Wikimedia Commons with a clear license record, detailed in Appendix~\ref{appendix:country-license-detail}.


\paragraph{Cultural Item Pool.}
\label{sec:culture-pool}
We select up to 5 cultural items for each category in each country, totaling 243 unique items. We extract only licensed images and their summarized item descriptions from Wikipedia or Wikimedia Commons. We further ensure their validity through a two-step quality check: (1) We cross-check the collected data against prior benchmarks \cite{romero2024cvqa,satar2025seeingculturebenchmarkvisual,winata-etal-2025-worldcuisines}; and (2) We employ internal quality control by conducting a blind, swapped peer review among all authors for every instance. Finally, each cultural item includes its name, country, description, and its corresponding image.

\subsection{Adversarial Pairing.}
\label{sec:pair-creation}

We employ 2 pairing methods following cultural proxies introduced by \citet{adilazuarda-etal-2024-towards}.
First, description-based pairing follows the semantic proxy, which captures cultural similarity based on the meaning of item descriptions.
Second, geographical-based pairing follows the demographic proxy, which reflects cultural closeness based on how near the countries are to each other.


For each item $x_i$, we find $x_j$ such that it forms two types of pairs: hard pair $\mathbf{p}_{i,j}^{\mathrm{hard}}$, representing the most semantically similar or geographically closest item pair, and the easy pair $\mathbf{p}_{i,j}^{\mathrm{easy}}$, representing the least semantically similar or geographically farthest item pair. Both pair's items always come from the same category.

\paragraph{Description.}


For description-based pairing, we measure semantic similarity between item descriptions. Each description is encoded into an embedding using the mE5 model \cite{wang2024multilinguale5textembeddings}, and the similarity score between two items is computed using cosine similarity:
$$S(x_i, x_j) = \frac{\mathbf{v}_i \cdot \mathbf{v}_j}{\|\mathbf{v}_i\| \, \|\mathbf{v}_j\|}.$$
The hardest and easiest pairs for a specific item $x_i$ are defined by finding the index $j$ that minimizes or maximizes this score, respectively, resulting in the pairs $\mathbf{p}_{i,j}^{\mathrm{hard}}$ and $\mathbf{p}_{i,j}^{\mathrm{easy}}$.

\begin{equation}
\begin{aligned}
\mathbf{p}_{i,j}^{\mathrm{hard}} &= (x_i, x_j) \quad : \quad j = \arg \min_{k \neq i} S(x_i, x_k), \\
\mathbf{p}_{i,j}^{\mathrm{easy}} &= (x_i, x_j) \quad : \quad j = \arg \max_{k \neq i} S(x_i, x_k).
\end{aligned}
\end{equation}

A higher similarity value indicates that the items share strong semantic overlap, making $\mathbf{p}_{i,j}^{\text{hard}}$ more likely to confuse the model. In this setting, an item $x_j$ serves as an adversarial example for $x_i$ when it closely resembles $x_i$ despite originating from a different culture. For example, \emph{lemper} from Indonesia and \emph{sushi} from Japan are both rice-based dishes often wrapped in natural leaves or seaweed. Although they belong to distinct culinary traditions, their textual and visual descriptions are closely aligned, making them more likely to confuse the model.

\paragraph{Geographical Distance.}
For geo-based pairing, we use geographic proximity between countries of origin. Each item $x_i$ is assigned the centroid coordinates $(\text{long}_i, \text{lat}_i)$ of its corresponding country, obtained using the \texttt{geopy} library.
We denote the distance between two items $x_i$ and $x_k$ as $D(x_i, x_k)$, which is computed using the Haversine formula.
\begin{equation}
\begin{aligned}
\mathbf{p}_{i,j}^{\mathrm{hard}} &= (x_i, x_j) : 
j = \arg \min_{k \neq i} D(x_i, x_k), \\
\mathbf{p}_{i,j}^{\mathrm{easy}} &= (x_i, x_j) : 
j = \arg \max_{k \neq i} D(x_i, x_k).
\end{aligned}
\end{equation}

The hard pair is expected to consist of items from geographically proximate regions that are more likely to share stylistic or historical influences despite national boundaries. For instance, \emph{batik} from Indonesia and \emph{songket} from Malaysia are produced in neighboring regions and exhibit overlapping textile motifs and weaving traditions. Such pairs may be challenging, making the model struggle to distinguish closely related regional cultures.

\begin{figure*}[t]
    \centering
    \begin{subfigure}[t]{0.7\linewidth} 
        \centering
        \includegraphics[width=\linewidth]{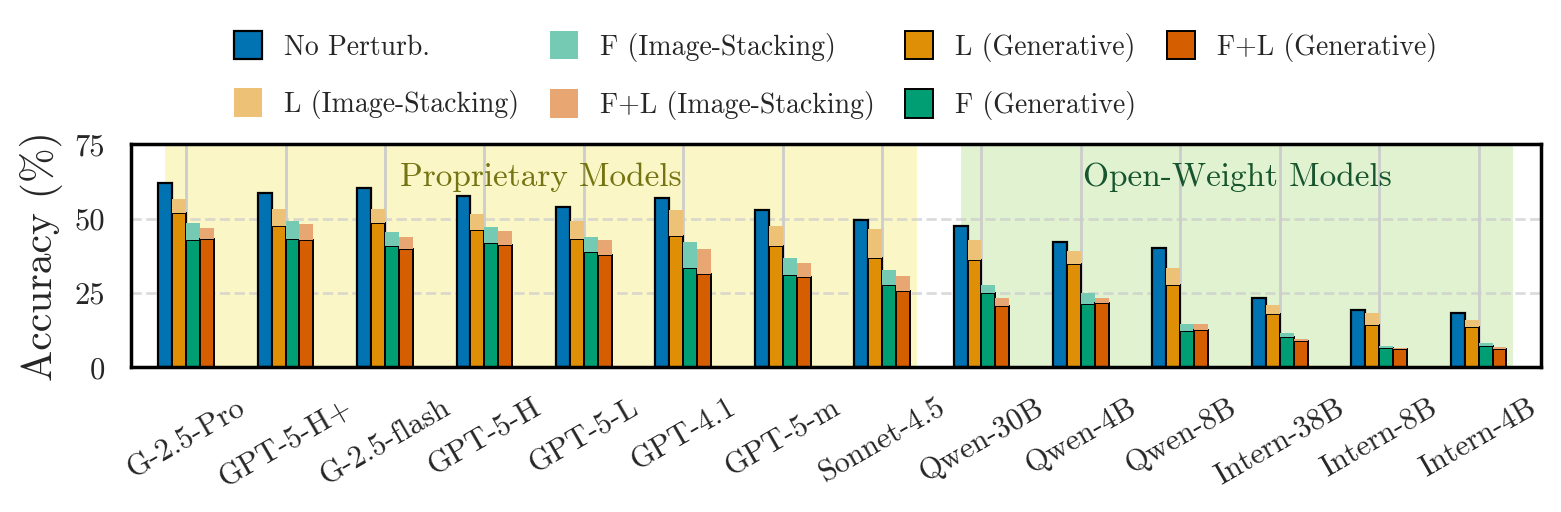}
        \caption{}
        \label{fig:context-by-model}
    \end{subfigure}%
    \begin{subfigure}[t]{0.158\linewidth} 
        \centering
        \includegraphics[width=\linewidth]{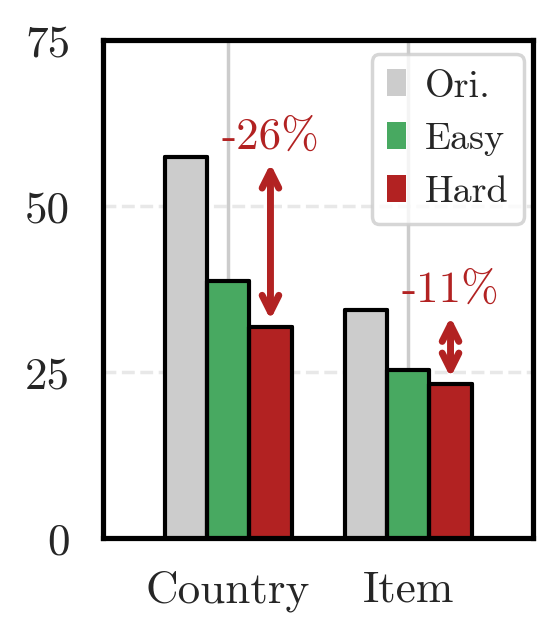}
        \caption{}
        \label{fig:difficulty-country-name}
    \end{subfigure}%
    \begin{subfigure}[t]{0.142\linewidth} 
        \centering
        \includegraphics[width=\linewidth]{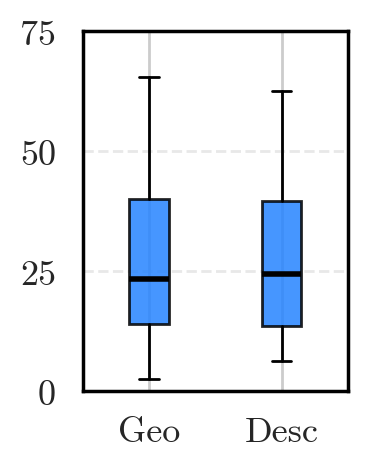}
        \caption{   }
        \label{fig:boxplot-geo-desc}
    \end{subfigure}
    \caption{Overall evaluation results in average accuracy of country \& cultural item prediction. Key trends: (a) Proprietary VLMs outperform open-weight variants, with generative perturbations being more adverse (especially with flags). (b) Predicting cultural item name is more challenging even from baseline case, though country accuracy drops are much larger in both difficulty levels. (c) Similar average performance of both pairing methods}

    \label{fig:experiment-figure}
\end{figure*}

\subsection{Image Perturbation}
We perturb the images using 2 approaches. First, we use an image-generation model, similar to the methodology used in prior work \cite{kimchi2025}. Second, we employ a simple image stacking perturbation to determine whether a simpler perturbation could already alter the model's accuracy. A handful of visual results produced via both approaches are provided in Appendix \ref{appendix:visual-example}.

\subsubsection{Image Stacking Perturbation}
\label{par:image_stacking}
We apply this perturbation by stacking the smaller adversarial image over the original cultural item image, which serves as a baseline adversarial attempt to assess models' robustness. We denote adversarial cue images as flag image ($\mathbf{I}_f$) and landmark image ($\mathbf{I}_l$). The size of the adversarial images is resized, denoted by the resizing operation $\mathcal{r}$, to maintain their original aspect ratio while ensuring neither dimension exceeds $20\%$ of the original item image dimension. Spatial placement is fixed and non-overlapping: $\mathbf{I}_f$ is consistently placed in the top-right corner ($\nearrow$), and $\mathbf{I}_l$ in the bottom-left corner ($\swarrow$), with a small offset from the edges. This stacking perturbation process, $\Phi$, is defined as a function of the original image $\mathbf{I}_{ori}$ and a set containing either or both of $\mathbf{I}_f$ and $\mathbf{I}_l$:

\begin{equation}
\label{eq:naive_perturbation}
\begin{aligned}
\mathbf{I}_{S} &= \Phi(\mathbf{I}_{ori}, \{\mathcal{r}(\mathbf{I}_f), \mathcal{r}(\mathbf{I}_l)\})
\end{aligned}
\end{equation}

\vspace{0.1em}
\subsubsection{Generative Perturbation}
\label{par:image_generation}
To curate an alternate perturbation case that is more immersive and naturally integrated, we use an image-generation model, specifically Gemini-2.5-Flash-Image \cite{nanobanana}. We denote this process as $\Psi$ resulting in a perturbed image ($\mathbf{I}_{G}$) as a function of the original item image ($\mathbf{I}_{ori}$), the adversarial cue images ($\mathbf{I}_{f}$, $\mathbf{I}_{l}$), a set containing either or both of $\mathbf{I}_f$ and $\mathbf{I}_l$, and a guiding prompt ($\mathbf{p}$):
\begin{align}
\mathbf{I}_{G} = \Psi(\mathbf{I}_{ori}, \{\mathbf{I}_{f}, \mathbf{I}_{l}\}, \mathbf{p}).
\end{align}
The strict prompt template ($\mathbf{p}$) and preserved inference hyperparameters are employed to best ensure spatial consistency, adversarial semantic guidance, and reproducibility. The details of which are outlined in Appendix \ref{appendix:image-gen}.

In attire-specific cases, our collected images often feature human body parts, posing a potential PII risk. Therefore, before inclusion in $\Psi$, we apply a preprocessing step where we use an image-generation model to isolate the clothing component. This process removes potential PII and preserves the cultural attire in a more neutral form. This enables subsequent adversarial perturbations without compromising privacy (see Appendix \ref{appendix:image-gen} for details).


We also conduct a manual quality check of the generative perturbed images. A random sample of 5\% from each category: cuisine, attire, and musical instrument was selected, as denoted previously with $\mathbf{I}_{G}$. Using the rubric defined in Appendix~\ref{appendix:rubrics-manual-eval}, the authors independently re-evaluated 130 sampled images, assigning scores on a 1–5 Likert scale. The generated images achieved an average score of 4.49, demonstrating the high quality of the perturbed outputs. However, quality varied across categories, with cuisine being the most realistic and attire the least.

\subsection{Evaluation Metrics}
\label{sec:metrics}
To evaluate the models' robustness and vulnerability to cultural perturbations, we use two metrics: one that measures general prediction accuracy across all categories, and another that measures the model's distractive behavior on incorrect country predictions.

\paragraph{Multi-Target Accuracy (Acc.).}
This metric measures substring match accuracy when each instance may have multiple ground-truth labels (e.g., alternative country or cultural item names). A prediction ($p_i$) is considered correct if the uncased prediction string is a substring of any uncased ground-truth string(s) ($\mathbf{G}_i$). The overall Accuracy (Acc.) is the average number of successful substring matches across all $N$ instances.

\begin{equation}
\text{Acc.} = \frac{1}{N} \sum_{i=1}^{N} \begin{cases}
1 & \text{if } \exists\, g \in \mathbf{G}_i \text{ s.t. } p_i \subseteq g \\
0 & \text{otherwise}
\end{cases}
\end{equation}


\paragraph{Model Distraction Likelihood (${D}_\mathcal{L}$).}
For each model, this metric measures the frequency with which an incorrect country prediction is directly deceived by the counterpart adversarial cue. This metric is essential for assessing model robustness because it quantifies the causal attribution of the error, rather than just counting general mistakes. For example, if a model is tasked with predicting Country A but is perturbed with content from Country B, a high score on this metric indicates that the model's error is a direct result of being swayed by the adversarial content from Country B, demonstrating a critical failure in visual grounding against geographic conflict.

\begin{equation}
\label{eq:distraction_rate}
\begin{aligned}
{D}_\mathcal{L} = P(\text{adv} \mid \text{wrong})
\end{aligned}
\end{equation}







  \begin{figure*}[t]
    \centering
    \begin{subfigure}[t]{0.65\linewidth} 
        \centering
        \includegraphics[width=\linewidth]{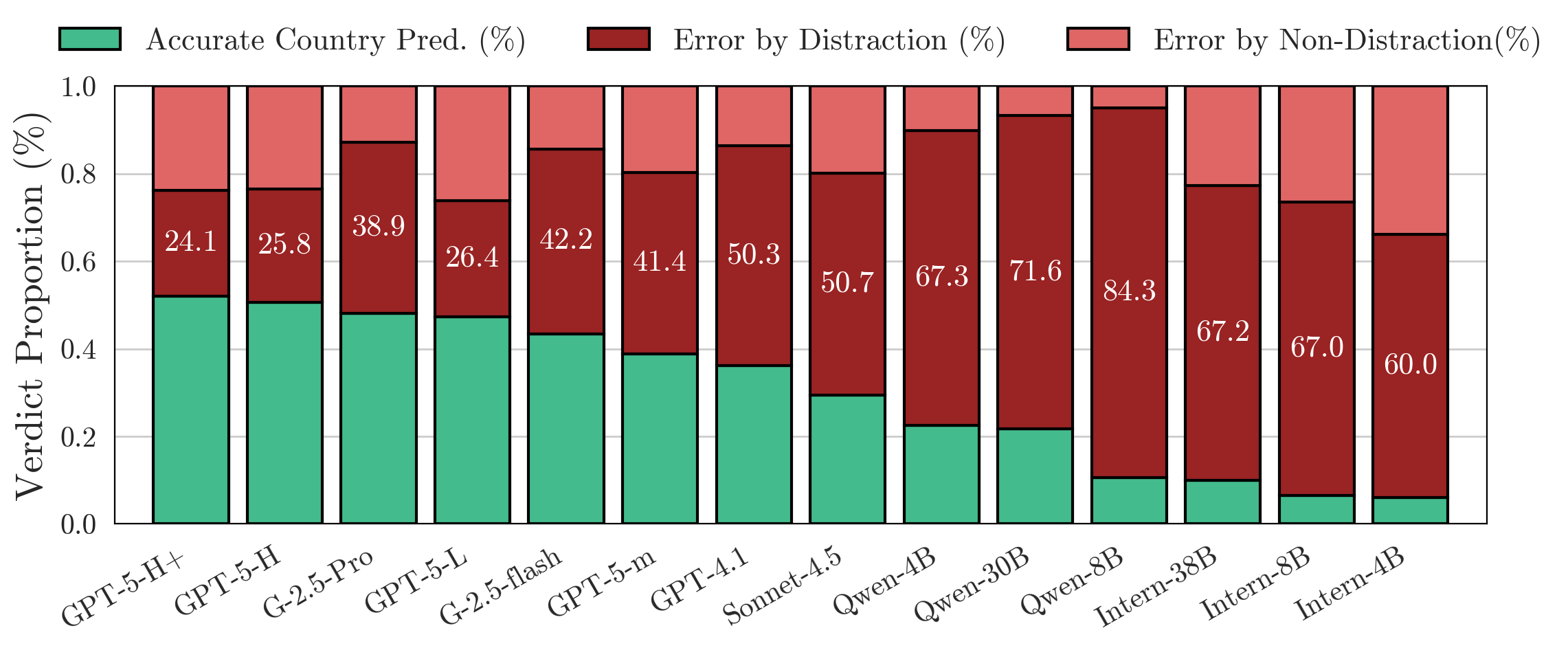}
        \caption{}
        \label{fig:distract-100}
    \end{subfigure}%
    \begin{subfigure}[t]{0.35\linewidth} 
        \centering
        \includegraphics[width=\linewidth]{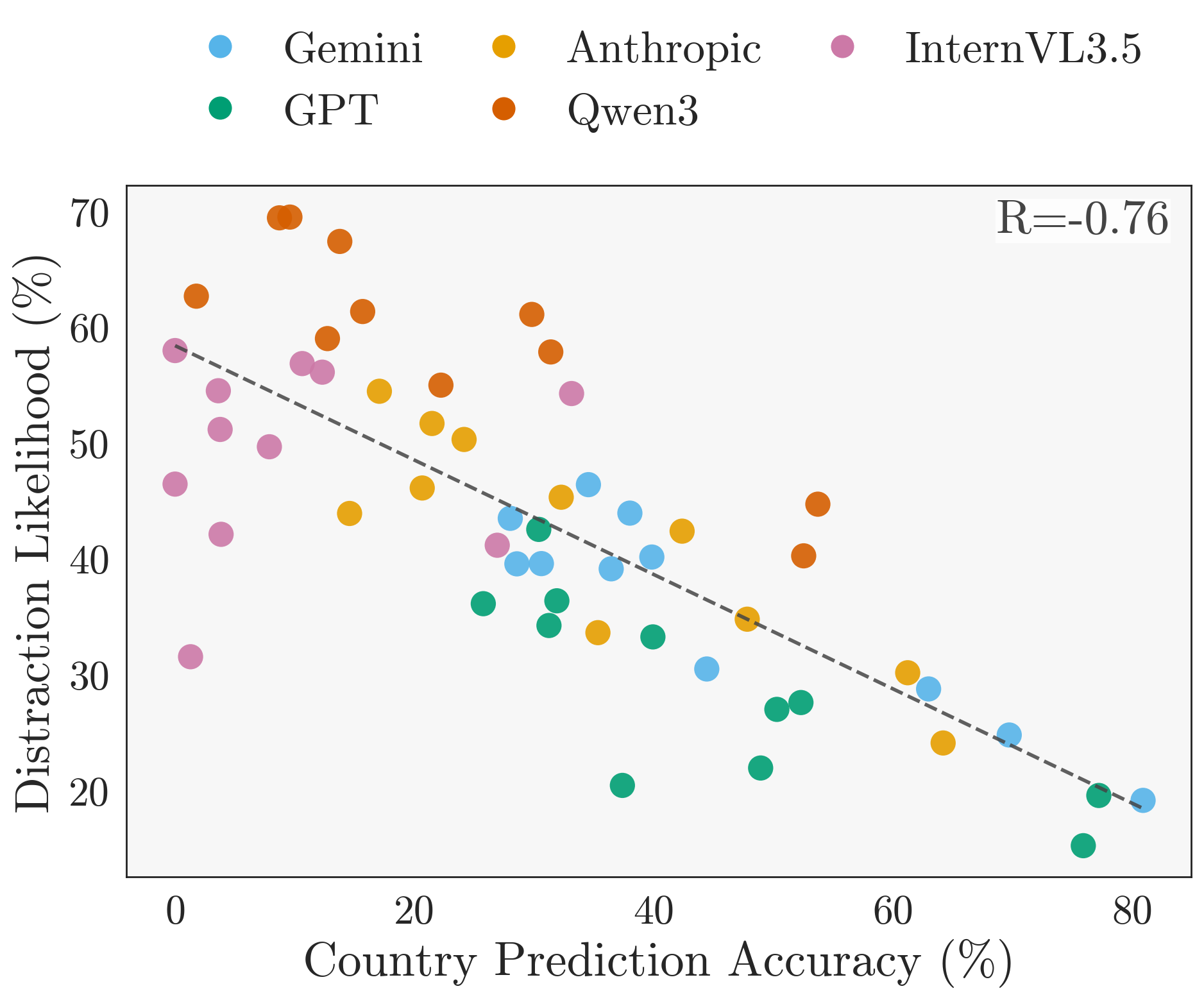}
        \caption{}
        \label{fig:distract-scatter}
    \end{subfigure}%
    \caption{The negative correlation between country prediction accuracy vs. distraction likelihood of the model in wrongly predicted cases. (a) The proportion of wrongly predicted countries across models increases along with the decrease of country prediction accuracy. (b) Across 11 different subregions for each VLM, the correlation of this relationship is also scoring at $-0.76$, suggesting a strong negative relation between the metrics.}
    \label{fig:distraction-analysis}
\end{figure*}

\section{Evaluation \& Results}

This section explains our evaluation methodology using \confusedtourist and presents the main results. We cover the model settings, the strict single-turn prompt protocol used for inference, as well as the core findings.

\subsection{Experiment Setup}

\paragraph{Model Selection.} We employ 14 model settings to evaluate current state-of-the-art VLMs. These settings are comprised of 8 proprietary variants, which spans across 3 model families (GPT, Gemini, Claude). We employ the other 6 open-weight settings that are part of 2 open-weight families, including Qwen3VL \cite{Qwen3-VL} and InternVL3.5 \cite{wang2025internvl35advancingopensourcemultimodal}.

\paragraph{Prompt.} All instances are being evaluated via a single inference call. VLMs are instructed to provide the answer for both the \textit{name} of the target cultural item and its \textit{country of origin}, as shown in the prompt below, where \texttt{\{category\}} is a choice between \texttt{attire, cuisine,} or \texttt{musical instrument}. We provide the detailed inference hyperparameters for each model in Appendix \ref{appendix:model-hyperparams}.

\noindent 
\fbox{%
    \parbox{\linewidth}{ 
        \raggedright 
        \scriptsize 
        \texttt{Observe the image and determine the original name of this \{category\} object, and the country from which this \{category\} originally comes. Return the original name of this \{category\} object first, followed by the country name}.
    }%
}\label{box:prompt}

\begin{figure*}[t!]
    \centering
    \begin{subfigure}[t]{0.60\linewidth}
        \centering
        \includegraphics[width=\linewidth]{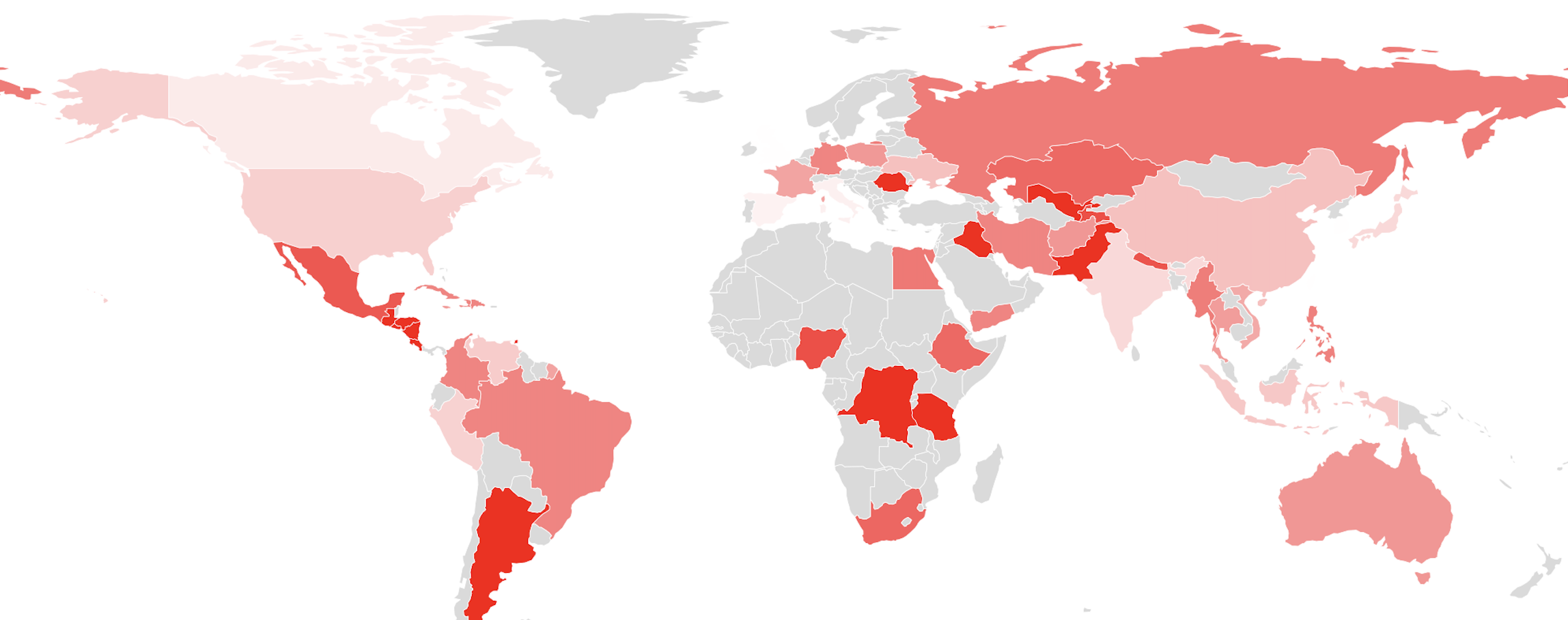}
        \caption{}
        \label{fig:placeholder}
    \end{subfigure}\hfill
    \begin{subfigure}[t]{0.35\linewidth}
        \centering
        \includegraphics[width=\linewidth]{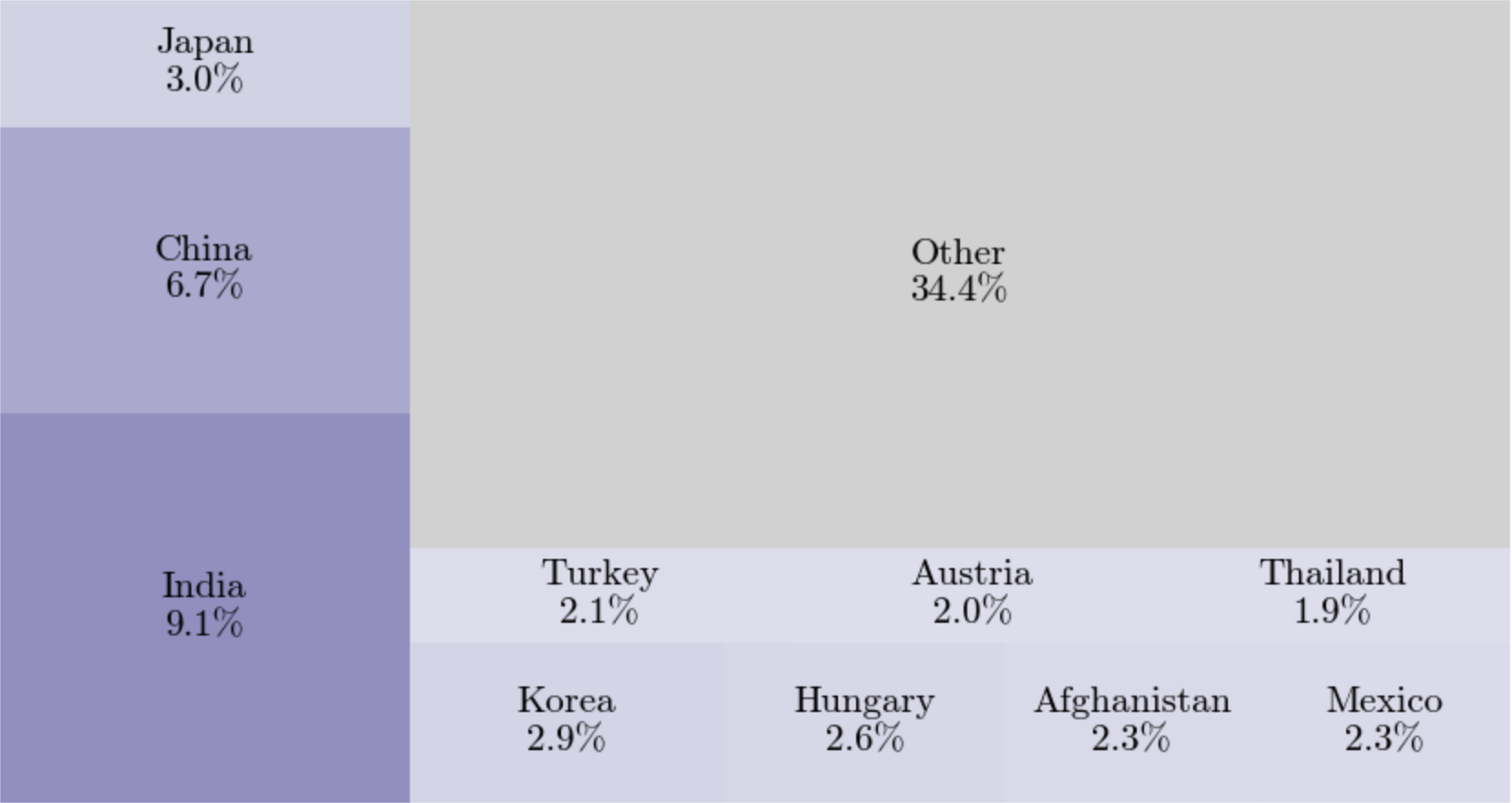}
        \caption{}
        \label{fig:TreemapBias}
    \end{subfigure}
    \caption{GPT-5 (High$^+$) results. (a) Global map of accuracy drop, computed as the ratio between performance difference and original score. (b) Distribution of predicted countries where the model is incorrect but does not follow the adversarial country}
    \label{fig:BiasFigure}
\end{figure*}

\subsection{Evaluation Result}
In this section, we outline the most apparent findings across all of our features. While the complete evaluation results can be observed in Table \ref{tab:complete-accuracy} and \ref{tab:complete-distraction}, we summarize the overall trend in major features in the points below by referring to Figure \ref{fig:experiment-figure}:

\begin{enumerate}[label=(\arabic*), itemsep=0pt, parsep=0pt, leftmargin=*, topsep=3pt]
    \item \textbf{Superior performance of proprietary VLMs}. Proprietary models outperform their open-weight variants, with the open-weight Qwen family scoring the closest to them, as depicted in Figure \ref{fig:context-by-model}. The difference in average prediction accuracy across the perturbation contexts (F, L, F+L) between the best performing GPT and Qwen3-VL variants is $22\%$ for country prediction and $27\%$ for item prediction. There is no significant benefit from the model's reasoning mode in either the baseline or in showing any improved resistance against adversarial context.
    \item \textbf{Flag object is the main perturbation driver.} In the image stacking setting, the presence of the flag caused a decline of up to 18.4\%, whereas applying the landmark perturbation resulted in a minor drop of only up to 6.9\%. Furthermore, the combination of their perturbation effect is observed to be no more than the sum of the individual parts, indicating that no emergent adversarial combination resulted from such combination.
    \item \textbf{Generative perturbation is more effective}, leading to an average performance worsening of 17.34\%, compared to only 8.43\% drop using image-stacking method. This validates that the generative (versus image-stacking) perturbation method is more successful in tricking the model. As observed in Figure \ref{fig:context-by-model}, the numbers are consistent in all cases when comparing shades within the same color in each perturbation context bar (F, L, F+L).
    \item \textbf{Proxy-based semantic and geographic cultural proxies remain relevant}. As observed in Figure \ref{fig:difficulty-country-name}, the difference in average accuracy drop is up to 26\% in the hardest cases (country) across cases. This signals that distinguishing morphologically similar items and items from geographically close cultures remains a challenge due to concept overlaps. Aligning with prior cultural proxy selections from \citet{adilazuarda-etal-2024-towards}, our pairing method also (description versus geographic-based) yields consistent results, as shown in Figure \ref{fig:boxplot-geo-desc}. This further suggests that both semantic and geographic cultural proxies remain relevant in multimodal multicultural evaluation.
\end{enumerate}

\section{Discussion \& Analysis}

In this section, we outline more fine-grained observations, where we go deeper discussing nature of observed visual distraction, existence of systematic country fallback biases, and the thought process of the VLMs in doing predictions.

\paragraph{Dominant Effect of Adversarial Cues in Prediction Errors.} 

We observe a clear inverse relationship: the lower a model's country prediction accuracy, the greater the tendency for its inaccurate predictions to drift directly toward our perturbation cues. Figure \ref{fig:distract-100} depicts this trend, highlighting that the decline in general accuracy is proportionate to the specific inaccuracy caused by distractive cues. This is further amplified by the significant negative correlation ($R=-0.76$) measured between these features across all cases, as shown in the scatter plot in Figure \ref{fig:distract-scatter}.

Amidst these findings, the Qwen family presents a compelling insight: despite maintaining a higher overall accuracy compared to InternVL3.5, it exhibits a higher Distraction Likelihood ($D_\mathcal{L}$). This finding suggests a pronounced their vulnerability to the adversarial cue specifically in error scenarios. We leverage this focused behavior—that their errors are less random—to select its 30B variant for deeper interpretability analysis, as it may help reducing the noise in the attention spread visualization.

\begin{figure}[b!]
    \centering
    \begin{subfigure}[t]{0.32\linewidth}
        \centering
        \includegraphics[width=\linewidth]{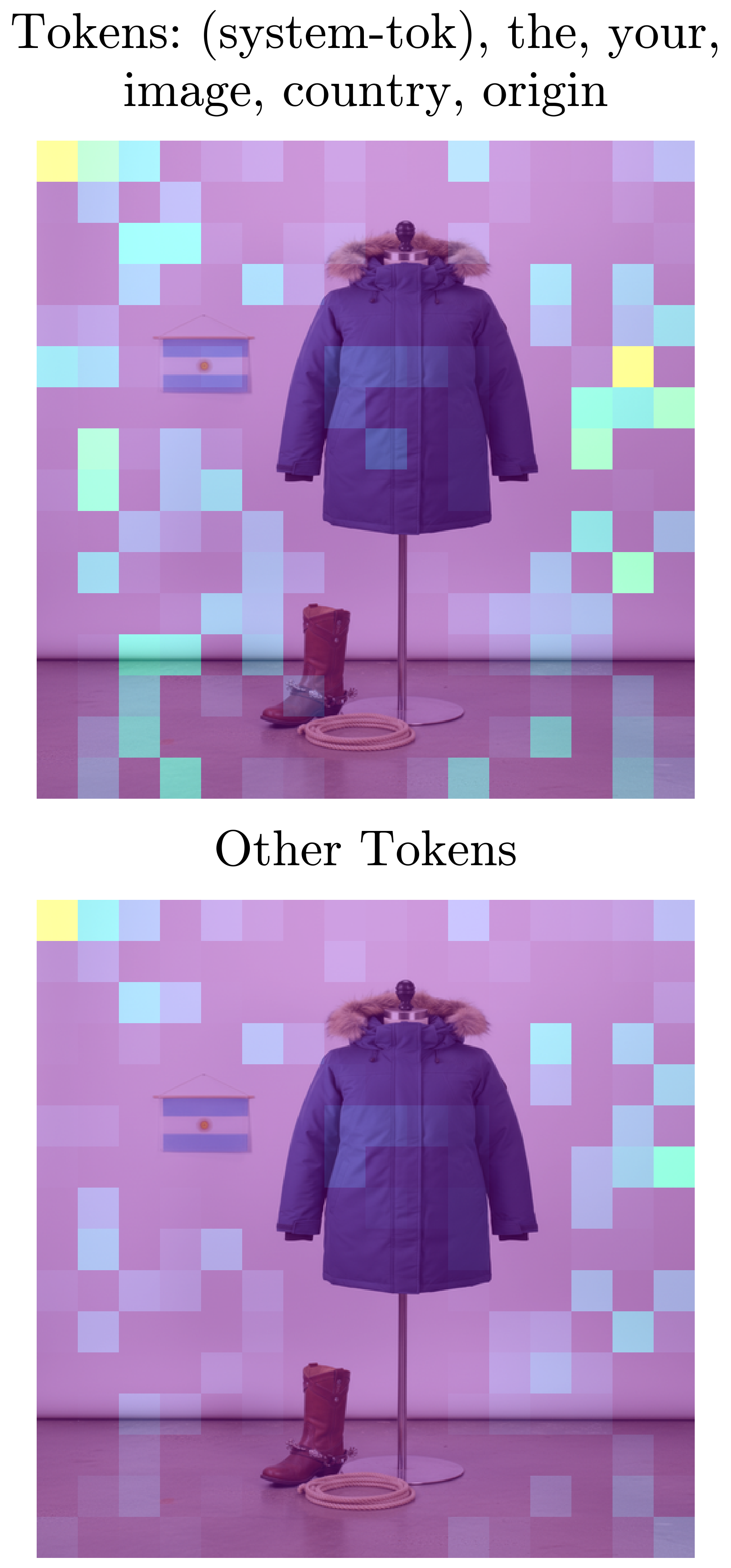}
        \caption{}
        \label{fig:token1}
    \end{subfigure}
    \hfill
    \begin{subfigure}[t]{0.32\linewidth}
        \centering
        \includegraphics[width=\linewidth]{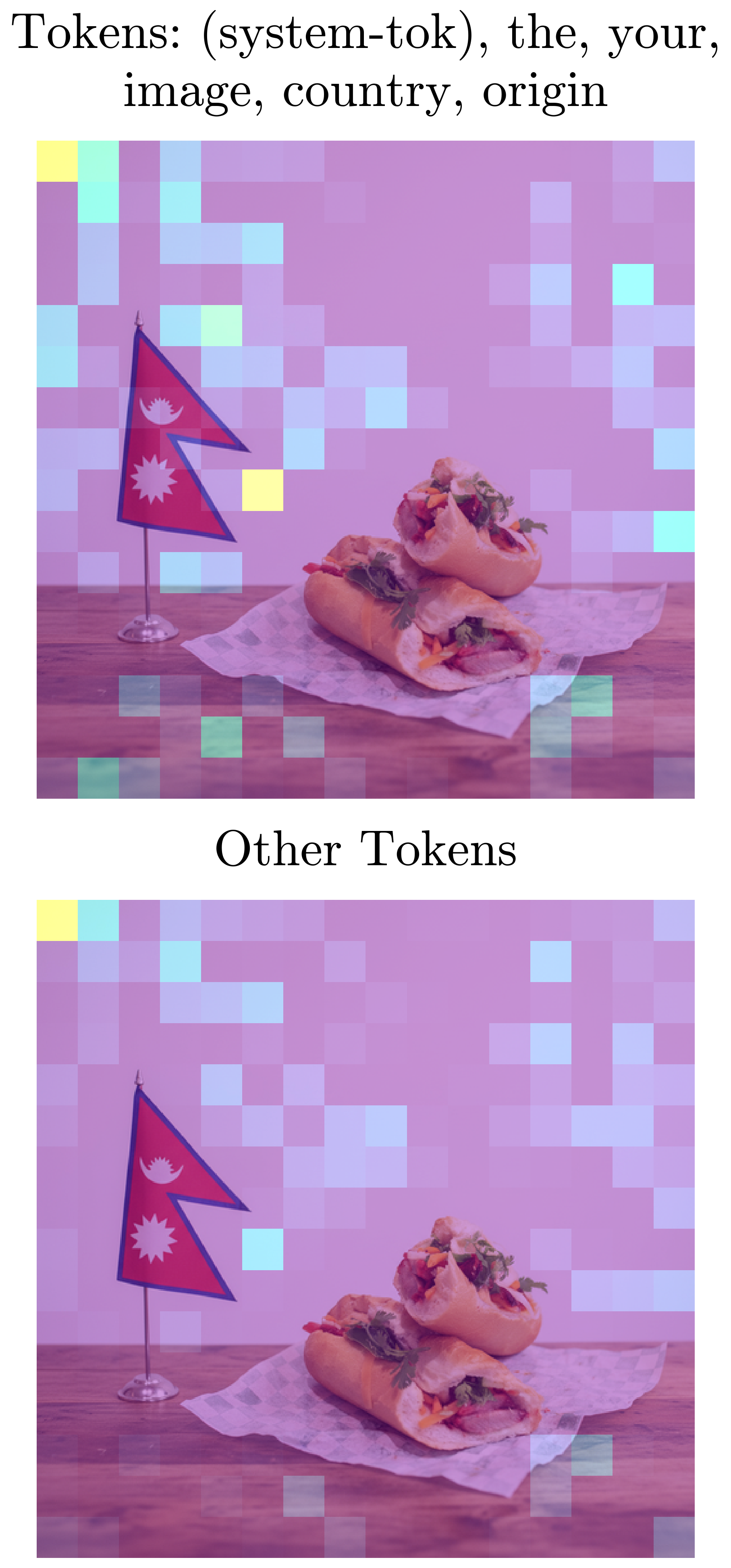}
        \caption{}
        \label{fig:token2}
    \end{subfigure}
    \hfill
    \begin{subfigure}[t]{0.32\linewidth}
        \centering
        \includegraphics[width=\linewidth]{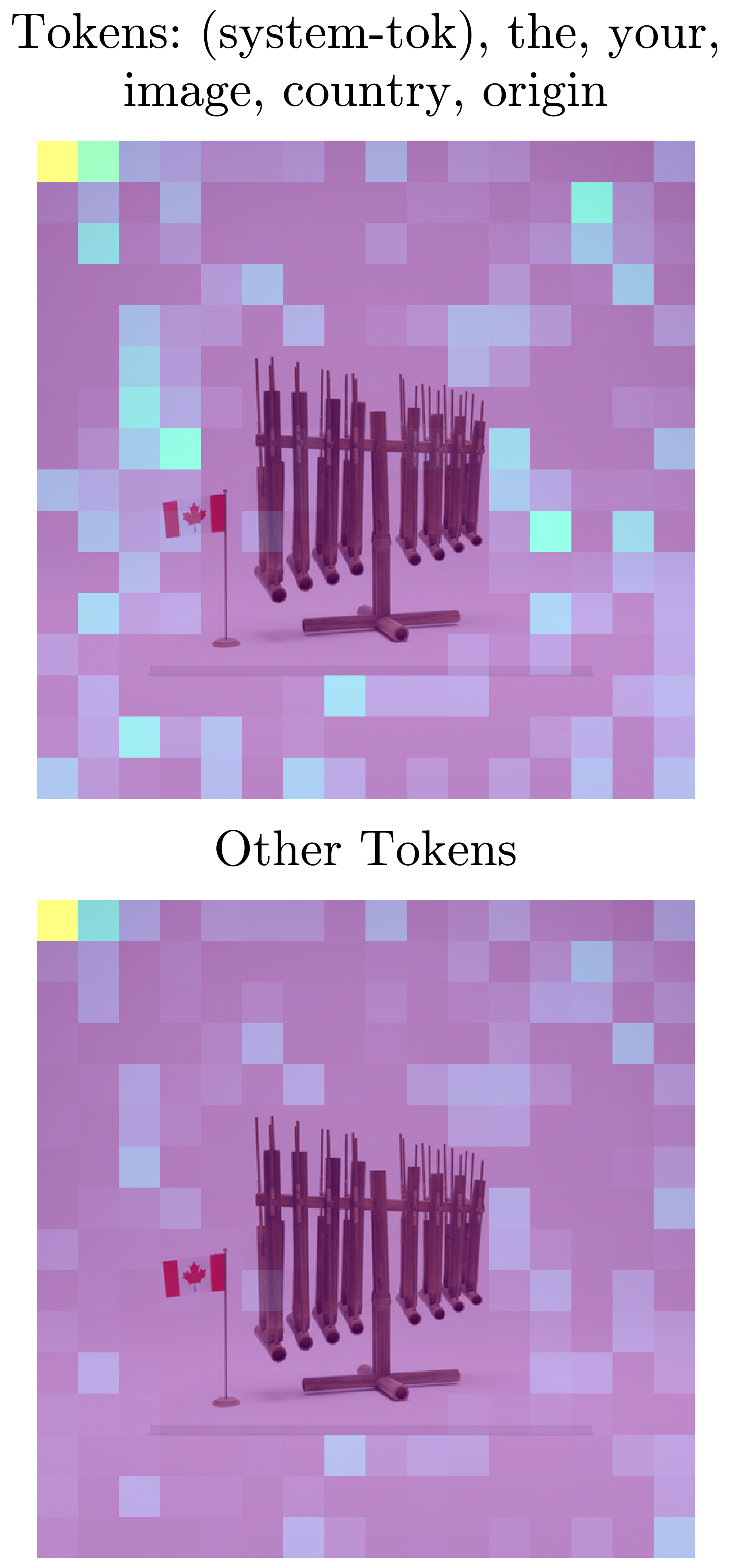}
        \caption{}
        \label{fig:token3}
    \end{subfigure}

    \caption{Attention heatmap analysis indicates that visual grounding primarily arises from a limited set of tokens. In (a) attire, (b) cuisine, and (c) musical-instrument culture items, tokens linked to system cues, geographic references, and category-specific terms dominate the model’s visual attention.}
    \label{fig:token-ablation}
\end{figure}

Our interpretability analysis revealed a critical finding: the model's attention focus on the image tokens was disproportionately driven by specific text components—namely, system tokens, geo-related tokens, and a subset of stopwords (as shown in Figure \ref{fig:token-ablation}). We further conducted an ablation study to investigate the effect of these tokens as detailed in Appendix \ref{appendix:token-ablation-study}. Critically, eliminating these suspected tokens resulted in two key improvements: a measurable increase in grounding accuracy and a pronounced shift in attention back to the intended image region. While this suggests an avenue for better prompt choice selection for our work, we still conclude that:

\begin{enumerate}
    \item VLMs tend to rely on easily interpretable visual cues (e.g., flags or other prominent context features), leading to failures in incorporating relevant knowledge or being overridden by familiar visual shortcuts.
    \item This behavior worsens due to reliance on specific text tokens, showing that even advanced VLMs can be unstable and highly sensitive to prompts—as seen in one failure case (Appendix \ref{appendix:token-ablation-study}, attire), where a prompt correction even turned a previously correct answer into a wrong one.
\end{enumerate}
These findings express the need for more globally aware models that remain stable under small, semantic-based input changes (i.e. in our case, specific geographic cues).

\paragraph{Specific Country Fallback Bias.}  
As shown in Figure~\ref{fig:BiasFigure}(a), GPT-5 (High$^+$) exhibits larger performance drops in several low-resource regions such as Africa, Latin America, and MENA. In contrast, supposedly low-resource regions like Southeast Asia (SEA) show relatively stable performance, possibly reflecting improved data coverage from recent community efforts such as SEACrowd~\cite{lovenia-etal-2024-seacrowd} and SEA-VL~\cite{cahyawijaya-etal-2025-crowdsource}.


To further investigate regional prediction tendencies, we analyze cases where the model’s predictions are incorrect but do not correspond to the adversarial country $\mathbf{C}_{adv}$. This scenario corresponds to the negation of $D_\mathcal{L}$ (Eq.~\ref{eq:distraction_rate}). As shown in Figure~\ref{fig:TreemapBias}, certain countries are predicted more frequently even in the absence of adversarial cues, suggesting a "fallback preference" toward specific regions.

For example, GPT-5 (High$^+$) tends to over-predict countries such as India (9.1\%), China (6.7\%), and Japan (3.0\%). Notably, India and China also appear in the top three most frequent incorrect predictions for all other models evaluated. This consistency suggests that image-text pairs from these cultures are more prevalent in the training distribution. We hypothesize that this over-representation may arise because these regions are common subjects of online imagery (e.g., as popular tourist destinations). 


\paragraph{Flag Cues Corrupt Model Reasoning Traces.}
We also analyze the model outputs by tracing the reasoning process of the Gemini model in sample cases where it produces incorrect answers. Our comparison considers several factors: perturbation type (flag, landmark, flag+landmark), category (cuisine, attire, musical instrument), and perturbation method (image-stacking vs. generated perturbation). From our observations, the model often exhibits uncertainty in the middle of its reasoning steps, with phrases resembling ``Am I correct? I am not sure''.

\definecolor{keyC}{RGB}{255,248,210} 
\newcommand{\kt}[1]{\colorbox{keyC}{\strut #1}}

\vspace{0.5em}
\noindent
\fbox{%
  \parbox{\linewidth}{%
    \raggedright
    \small

    \textbf{Flag perturbation — Reasoning}\\
    \textit{“The style of this garment screams West or Central African influence to me. The wide sleeves, the embroidery... I’m leaning heavily towards the \kt{DRC} \kt{(from the flag)}. I need to find the specific name for this type of clothing in the DRC.”}\\[1.9em]

    \textbf{Landmark perturbation — Reasoning}\\
    \textit{“I’m thinking thobe, dishdasha, kandura, maybe even jalabiya. Could even be a simplified caftan, given the Turkish context. Since the prompt is about the \kt{origin} (not where it’s \kt{displayed}), I need to be more precise.”}\\[1.9em]

    \textbf{Flag + Landmark perturbation — Reasoning}\\
    \textit{“I see a white blouse with red and black embroidery on a mannequin. There’s a flag visible — a vertical tricolor of blue, yellow, and red \kt{(Romania)}. The background has that massive building — the \kt{Palace of the Parliament (Bucharest)}. Definitely Romania.”}
  }%
}
\vspace{0.5em}

Among these factors, the most distinct differences appear in the perturbation type. The model typically focuses on the flag first before attending to other relevant aspects of the image. Landmark perturbations are generally less effective, as the model sometimes concludes that ``there is no connection between the landmark and the object'', leading it to refocus on the primary object instead. However, in flag+landmark cases, the model tends to perceive a strong correlation between the flag and the landmark, and thus shifts its reasoning toward these contextual cues rather than the main object itself.

In contrast, the category and perturbation method do not appear to introduce substantial differences, as the model’s reasoning predominantly centers around the flag across most cases.

\section{Conclusion}

We introduced \confusedtourist, a novel cultural adversarial robustness suite designed to evaluate VLMs' ability to accurately identify cultural items in adversarially-induced images. Our experiments reveal a critical vulnerability: all state-of-the-art VLMs experience a substantial accuracy drop under simple image stacking, which becomes even more severe under generative perturbations. We found these geographical-induced perturbations consistently cause disruption across all cases, a vulnerability that is more prominent with the presence of a flag, which consistently exhibits the model's grounding bias toward the adversarial cue. Further analysis shows that as model accuracy decreases, the models become increasingly distracted by these perturbations, focusing more on the distractor cues than on the cultural item itself, with the reasoning traces supporting this observation. Overall, this work highlights a critical challenge: VLMs must develop greater cultural robustness to achieve reliable multimodal understanding across diverse cultural contexts.

{
    \small
    \bibliographystyle{ieeenat_fullname}
    \bibliography{main}
}

\clearpage
\setcounter{page}{1}
\maketitlesupplementary

\begin{table*}[t]
    \centering
    \small
    \caption{Geographic Breakdown of Countries in the Dataset by Region and Sub-Region, with Regional Totals.}
    \label{tab:geo_breakdown_list_detailed_twocol_column_final}
    \begin{tabular}{@{} l l p{6cm} r @{}}
        \toprule
        \textbf{Region} & \textbf{Sub-Region} & \textbf{Countries} & \textbf{Region Total} \\
        \midrule
        \multirow{2}{*}{Africa} & Sub-Saharan Africa & DR Congo, Ethiopia, Nigeria, South Africa, Tanzania & \multirow{2}{*}{\textbf{10}} \\
         & The Middle East \& North Africa (MENA) & Egypt, Iran, Iraq, Turkey, Yemen &  \\
        \midrule
        \multirow{2}{*}{Americas} & North America & Canada, Cuba, Guatemala, Mexico, United States & \multirow{2}{*}{\textbf{10}} \\
         & South America & Argentina, Brazil, Colombia, Peru, Venezuela &  \\
        \midrule
        \multirow{4}{*}{Asia} & Central Asia & Kazakhstan, Kyrgyzstan, Tajikistan, Turkmenistan, Uzbekistan & \multirow{4}{*}{\textbf{20}} \\
         & East Asia & China, Hong Kong, Japan, South Korea, Taiwan &  \\
         & South Asia & Afghanistan, Bangladesh, India, Nepal, Pakistan &  \\
         & Southeast Asia & Indonesia, Myanmar, Philippines, Thailand, Vietnam &  \\
        \midrule
        \multirow{2}{*}{Europe} & Eastern Europe & Czechia, Poland, Romania, Russia, Ukraine & \multirow{2}{*}{\textbf{10}} \\
         & Western Europe & France, Germany, Italy, Spain, United Kingdom &  \\
        \midrule
        \multirow{1}{*}{Oceania} & Oceania & Australia, Fiji, New Zealand & \multirow{1}{*}{\textbf{3}} \\
        \bottomrule
        \multicolumn{3}{r}{\textbf{Grand Total}} & \textbf{53} \\
    \end{tabular}
\end{table*}


\section{Country \& Image License Details}
\label{appendix:country-license-detail}

All data used in this work are sourced from Wikimedia Commons. Each instance whether it represents a flag, an item (e.g., attire, cuisine, or musical instrument), or a landmark, we include its corresponding license information, as shown in Table~\ref{tab:dataset_attributes}. We strictly include only media that are permitted for research use and redistribution. 

\begin{table*}[t]
\centering
\caption{Dataset attribute descriptions and examples. License information is provided as dedicated rows for each URL-bearing field.}
\begin{tabular}{@{}p{2.6cm}p{1.8cm}p{7.7cm}p{3.2cm}@{}}
\toprule
\textbf{Attribute} & \textbf{Type} & \textbf{Description} & \textbf{Example} \\
\midrule
id & int32 & Unique identifier for each record. & 101 \\
image & image & Main image associated with the item. & (Image file) \\
item & string & Name or title of the item (e.g., dish or object). & Eiffel Tower \\
origin\_country & string & Country of origin for the item. & France \\
adversarial\_country & string & Country used for adversarial comparison or challenge. & Germany \\
category & string & Category or type of the item. & Landmark \\
difficulty & string & Difficulty level of the task or challenge. & Medium \\
perturb\_method & string & Method used to perturb or modify the image/data. & Gaussian Noise \\
landmark\_name & string & Name of the landmark or key feature in the image. & Eiffel Tower \\
perturb\_context & string & Context of the perturbation or transformation applied. & Added clouds and reduced brightness. \\
pair\_method & string & Method used to pair original and perturbed images. & Nearest neighbor similarity \\
item\_url & string & URL linking to more information about the item. & \url{https://en.wikipedia.org/wiki/Eiffel_Tower} \\
item\_url\_license & string & License for the content at \texttt{item\_url}. & CC BY-SA 3.0 \\
flag\_url & string & URL to the flag image of the associated country. & \url{https://upload.wikimedia.org/wikipedia/en/c/c3/Flag_of_France.svg} \\
flag\_url\_license & string & License for the content at \texttt{flag\_url}. & Public Domain \\
landmark\_url & string & URL linking to the landmark’s reference/source image. & \url{https://commons.wikimedia.org/wiki/File:Eiffel_Tower_Paris.jpg} \\
landmark\_url\_license & string & License for the content at \texttt{landmark\_url}. & CC BY-SA 3.0 \\
\bottomrule
\end{tabular}
\label{tab:dataset_attributes}
\end{table*}

\section{Evaluation on Prior Benchmarks}
\label{appendix:eval-prior}

We benchmark GPT-5 as the new upper bound model for prior benchmarks namely CVQA \cite{romero2024cvqa} and WorldCuisines \cite{winata-etal-2025-worldcuisines} to assess improvements over previously evaluated models (e.g., GPT-4), which represent the strongest reported results to date.

For CVQA, we evaluated the model in the location-agnostic setting using the prompt format
“Question: {question} Options: {options} Short Answer:”
without including any country information. This setting yielded a score of 74.7\%, whereas the best result reported in the paper was 48.7\% (Instruct BLIP \cite{10.5555/3666122.3668264}).

For WorldCuisines, we evaluated both Task 1 (cuisine identification) and Task 2 (country prediction) in the multiple-choice (MCQ) setting. We followed the custom prompt provided for each question in the benchmark. GPT5 model achieved 92.7\% for task 1, and 78.4\% for task 2, compared to the best reported performance of 88.4\% and 66.52\% (GPT-4o)

\section{Image Generation Prompt \& Model Settings}
\label{appendix:image-gen}

The image generation process utilizes the \texttt{gemini-2.5-flash-image} model with fixed configuration settings ($\text{temperature}=0.0$, $\text{aspect\_ratio}=1:1$, $\text{max\_output\_tokens}=648$). The methodology aims to test model robustness by generating a photorealistic image where a traditional cultural item (\texttt{\{item\_name\}} from \texttt{\{origin\_country\}}) is integrated into the environment of an \texttt{\{adversarial\_country\}}. The prompt template is constrained to curate a central object placement (occupying $\ge 1/3$ area). The environmental blend is achieved by inserting a \texttt{\{background\_instruction\}} corresponding to the test mode (\texttt{flag}, \texttt{landmark}, or \texttt{flag\_landmark}) and a supporting instruction based on the object \texttt{\{category\}} (e.g., table for Cuisine).

The complete image generation prompt text is structured as follows, with placeholders filled by the experimental variables:

\noindent
\fbox{%
    \parbox{\linewidth}{%
        \raggedright
        \scriptsize
        \texttt{Context: I am providing you a traditional \{category\} from \{origin\_country\} called \{item\_name\}.
        Generate a photorealistic image of this \{category\} as if it is situated in \{adversarial\_country\}, blending it naturally into the new environment.}

        \vspace{0.5em}

        \texttt{Image generation rules:}
        \begin{itemize}
            \item \texttt{The \{category\} object should be placed in the center of the image, taking up at least 1/3 of the total image area to ensure it's prominently featured and clearly visible.}
            \item \texttt{\{background\_instruction\}}
            \item \texttt{Maintain the originality, completeness, and detail of the \{category\} object itself.}
            \item \texttt{Integrate the \{category\} and the new scenery by adding a relevant supporting object to make the scene natural and cohesive (\{supporting\_bg\_instruction\}).}
            \item \texttt{IMPORTANT: Do not include any people, human figures, or human silhouettes in the background or anywhere in the image. Focus only on the \{category\} object and the environmental scenery.}
            \item \texttt{The image should be photorealistic, with the object and environment being blended seamlessly so it looks completely natural and not artificially composed.}
            \item \texttt{Avoid any obvious compositing artifacts, unnatural lighting, or perspective mismatches between the \{category\} and its surroundings.}
            \item \texttt{Ensure consistent lighting, shadows, and color grading across the entire image to enhance realism.}
        \end{itemize}
    }%
}\label{box:imageprompt}

\section{Manual Evaluation Rubrics}
\label{appendix:rubrics-manual-eval}

We define the rubrics for evaluating the quality of the model-assisted perturbation below. Worth noting that we use the same number of samples per category, so no category is left behind. Based on that rubric, the generated image perturbation is 4.49, which reflects the high quality of the perturbed outputs.

\begin{itemize}
    \item \textbf{5 (Excellent)}. Perturbation is highly natural and seamless. The added element (flag, landmark, or both) blends perfectly with lighting, perspective, and style. No visible artifacts, mismatched colors, or unnatural edges. The image appears authentic and coherent.
    \item \textbf{4 (Good)}. Perturbation is mostly natural. Minor inconsistencies in lighting, scale, or blending are noticeable on close inspection but do not significantly harm realism. Overall, the image looks believable.
    \item \textbf{3 (Fair)}. Perturbation is moderately convincing. Integration issues such as slight misalignment, contrast mismatch, or unrealistic positioning are visible. The edit is understandable but clearly artificial.
    \item \textbf{2 (Poor)}. Perturbation is visibly artificial. Clear signs of editing, such as wrong lighting, perspective errors, or poor blending. The added element does not integrate well with the original image.
    \item \textbf{1 (Very Poor)}. Perturbation is unnatural or of low quality. Severe mismatches in scale, lighting, or realism. The added element appears pasted-on, distorted, or contextually incoherent. The image looks fake or broken.
\end{itemize}

\section{Image Stacking Perturbation Pseudocode}
\label{appendix:image-stack}

This section provides the detailed pseudocode of procedure described in Section~\ref{par:image_stacking}. The algorithm outlines how adversarial cues—flags and landmarks—are overlaid onto original cultural item images to generate perturbed samples. Each perturbation mode (\textit{flag}, \textit{landmark}, \textit{flag\_landmark}) follows a controlled compositing process defined in Algorithm~\ref{alg:naive_perturbation}.

\begin{algorithm}[hbtp]
\small
\caption{Naive Image Stacking Perturbation}
\label{alg:naive_perturbation}
\KwIn{$\mathcal{D}_{pair}, \mathcal{D}_{geo}, m \in \{\text{flag}, \text{landmark}, \text{flag\_landmark}\}, \mathcal{A}$}
\KwOut{$\mathcal{I}_S, \mathcal{L}$}

\ForEach{$(i, d_i) \in \mathcal{D}_{pair}$}{
  \ForEach{$a \in \mathcal{A}$}{
    $c_{adv} \leftarrow \psi(d_i, a)$; \quad $\mathbf{g}_{adv} \leftarrow \gamma(\mathcal{D}_{geo}, c_{adv})$ \;
    $(\mathbf{I}_f, \mathbf{I}_l) \leftarrow \eta(\mathbf{g}_{adv})$; \quad $(\mathbf{I}_{ori}, \mathbf{I}_f, \mathbf{I}_l) \leftarrow \nu(\mathbf{I}_{ori}, \mathbf{I}_f, \mathbf{I}_l, m)$\;
    
    \If{$m \in \{\text{flag}, \text{flag\_landmark}\}$}{
        $(w_f,h_f) \leftarrow \mathcal{r}(\mathbf{I}_f; \tfrac{1}{5}\mathbf{I}_{ori})$; place $\mathbf{I}_f$ at $\nearrow$\;
    }
    \If{$m \in \{\text{landmark}, \text{flag\_landmark}\}$}{
        $(w_l,h_l) \leftarrow \mathcal{r}(\mathbf{I}_l; \tfrac{1}{5}\mathbf{I}_{ori})$; place $\mathbf{I}_l$ at $\swarrow$\;
    }

    $\mathbf{I}_S \leftarrow \Phi(\mathbf{I}_{ori}, \{\mathbf{I}_f, \mathbf{I}_l\})$; \quad $\mathcal{L} \leftarrow \mathcal{L} \cup \{\mathbf{I}_S\}$\;
  }
}
\Return{$\mathcal{I}_S, \mathcal{L}$}
\end{algorithm}

\section{Complete Evaluation Results}
\label{appendix:eval-results}

\begin{table*}[!t]
\centering
\caption{Performance drop comparison across models under different perturbation settings. Columns indicate multilabel exact matching accuracy (\%) for naive and AI-based perturbation methods across landmark (L), flag (F), and both (L+F) perturbation context infusion. Difficulty levels correspond to distance for Geo and semantic similarity for Desc as discussed in Sec. \ref{sec:pair-creation}. Detailed inference hyperparameters and settings for each model are outlined in Appendix \ref{appendix:model-hyperparams}.}
\label{tab:complete-accuracy}
\resizebox{\textwidth}{!}{
\begin{tabular}{llcccccccccccccc}
\toprule
\multirow{3}{*}{} & \multirow{3}{*}{} & \multirow{3}{1cm}{\centering Without\\ Perturb.}
& \multicolumn{6}{c}{Naive}
& \multicolumn{6}{c}{AI} \\
\cmidrule(lr){4-9} \cmidrule(lr){10-15}
& & & \multicolumn{3}{c}{Geo} & \multicolumn{3}{c}{Desc}
& \multicolumn{3}{c}{Geo} & \multicolumn{3}{c}{Desc} \\
\cmidrule(lr){4-6} \cmidrule(lr){7-9}
\cmidrule(lr){10-12} \cmidrule(lr){13-15}
 & & & L & F & L+F & L & F & L+F & L & F & L+F & L & F & L+F \\
\midrule
\textit{Difficulty} & & & Easy/Hard & Easy/Hard & Easy/Hard & Easy/Hard & Easy/Hard & Easy/Hard & Easy/Hard & Easy/Hard & Easy/Hard & Easy/Hard & Easy/Hard & Easy/Hard \\
\midrule
\textbf{Proprietary} \\
\midrule
GPT-5 (High$^+$) & Country & 66.7 & \rgscore{-0.01}{65.9}/\rgscore{-0.09}{61.6} & \rgscore{-0.07}{62.6}/\rgscore{-0.30}{49.3} & \rgscore{-0.09}{61.6}/\rgscore{-0.31}{48.8} & \rgscore{-0.03}{64.7}/\rgscore{-0.13}{59.2} & \rgscore{-0.16}{57.4}/\rgscore{-0.21}{54.5} & \rgscore{-0.15}{57.9}/\rgscore{-0.18}{56.4} & \rgscore{-0.29}{49.8}/\rgscore{-0.29}{49.8} & \rgscore{-0.34}{46.9}/\rgscore{-0.44}{41.2} & \rgscore{-0.36}{46.0}/\rgscore{-0.41}{42.7} & \rgscore{-0.33}{47.7}/\rgscore{-0.43}{41.7} & \rgscore{-0.40}{43.4}/\rgscore{-0.55}{34.6} & \rgscore{-0.38}{44.7}/\rgscore{-0.51}{37.0} \\
 & Item & 57.6 & \rgscore{-0.07}{54.5}/\rgscore{-0.10}{53.1} & \rgscore{-0.08}{54.0}/\rgscore{-0.21}{47.9} & \rgscore{-0.14}{51.2}/\rgscore{-0.18}{49.3} & \rgscore{-0.06}{54.9}/\rgscore{-0.18}{49.3} & \rgscore{-0.12}{51.9}/\rgscore{-0.16}{50.2} & \rgscore{-0.12}{51.9}/\rgscore{-0.23}{46.9} & \rgscore{-0.33}{42.2}/\rgscore{-0.17}{49.8} & \rgscore{-0.36}{40.8}/\rgscore{-0.39}{39.8} & \rgscore{-0.32}{42.7}/\rgscore{-0.38}{40.3} & \rgscore{-0.26}{45.5}/\rgscore{-0.44}{37.4} & \rgscore{-0.37}{40.4}/\rgscore{-0.51}{34.1} & \rgscore{-0.46}{36.2}/\rgscore{-0.53}{33.2} \\
GPT-5 (High) & Country & 66.7 & \rgscore{-0.01}{66.4}/\rgscore{-0.06}{63.0} & \rgscore{-0.04}{64.5}/\rgscore{-0.29}{49.8} & \rgscore{-0.17}{56.9}/\rgscore{-0.28}{50.2} & \rgscore{-0.08}{62.1}/\rgscore{-0.13}{59.2} & \rgscore{-0.11}{60.4}/\rgscore{-0.24}{52.6} & \rgscore{-0.12}{59.6}/\rgscore{-0.26}{51.7} & \rgscore{-0.23}{53.1}/\rgscore{-0.29}{49.8} & \rgscore{-0.34}{46.9}/\rgscore{-0.49}{37.9} & \rgscore{-0.36}{46.0}/\rgscore{-0.43}{41.7} & \rgscore{-0.32}{48.1}/\rgscore{-0.45}{40.3} & \rgscore{-0.40}{43.4}/\rgscore{-0.56}{34.1} & \rgscore{-0.38}{44.7}/\rgscore{-0.57}{33.6} \\
 & Item & 56.4 & \rgscore{-0.13}{50.7}/\rgscore{-0.15}{49.8} & \rgscore{-0.13}{50.7}/\rgscore{-0.26}{44.5} & \rgscore{-0.22}{46.4}/\rgscore{-0.27}{44.1} & \rgscore{-0.11}{51.5}/\rgscore{-0.26}{44.5} & \rgscore{-0.14}{50.2}/\rgscore{-0.37}{39.8} & \rgscore{-0.18}{48.1}/\rgscore{-0.31}{42.2} & \rgscore{-0.36}{40.3}/\rgscore{-0.25}{45.0} & \rgscore{-0.38}{39.3}/\rgscore{-0.42}{37.4} & \rgscore{-0.38}{39.3}/\rgscore{-0.41}{37.9} & \rgscore{-0.32}{42.1}/\rgscore{-0.49}{34.1} & \rgscore{-0.36}{40.0}/\rgscore{-0.58}{30.3} & \rgscore{-0.42}{37.4}/\rgscore{-0.55}{31.8} \\
GPT-5 (Low) & Country & 63.8 & \rgscore{-0.06}{60.7}/\rgscore{-0.12}{57.3} & \rgscore{-0.11}{57.8}/\rgscore{-0.35}{44.5} & \rgscore{-0.19}{53.1}/\rgscore{-0.29}{47.9} & \rgscore{0.00}{63.8}/\rgscore{-0.13}{56.4} & \rgscore{-0.14}{56.2}/\rgscore{-0.26}{49.3} & \rgscore{-0.14}{56.2}/\rgscore{-0.25}{49.8} & \rgscore{-0.31}{46.4}/\rgscore{-0.29}{47.4} & \rgscore{-0.38}{42.7}/\rgscore{-0.46}{37.9} & \rgscore{-0.42}{40.3}/\rgscore{-0.46}{37.9} & \rgscore{-0.30}{47.2}/\rgscore{-0.50}{36.0} & \rgscore{-0.43}{39.6}/\rgscore{-0.57}{32.2} & \rgscore{-0.38}{42.6}/\rgscore{-0.61}{29.9} \\
 & Item & 50.2 & \rgscore{-0.06}{47.9}/\rgscore{-0.15}{44.1} & \rgscore{-0.09}{46.4}/\rgscore{-0.20}{42.2} & \rgscore{-0.18}{43.1}/\rgscore{-0.22}{41.2} & \rgscore{-0.12}{45.5}/\rgscore{-0.19}{42.7} & \rgscore{-0.14}{44.7}/\rgscore{-0.37}{35.5} & \rgscore{-0.17}{43.4}/\rgscore{-0.28}{38.9} & \rgscore{-0.34}{36.5}/\rgscore{-0.23}{40.8} & \rgscore{-0.34}{36.5}/\rgscore{-0.38}{35.1} & \rgscore{-0.34}{36.5}/\rgscore{-0.37}{35.5} & \rgscore{-0.34}{36.6}/\rgscore{-0.50}{30.3} & \rgscore{-0.42}{33.2}/\rgscore{-0.53}{28.9} & \rgscore{-0.32}{37.4}/\rgscore{-0.63}{25.1} \\
GPT-5 (Minimal) & Country & 63.4 & \rgscore{-0.01}{63.0}/\rgscore{-0.10}{57.8} & \rgscore{-0.21}{51.7}/\rgscore{-0.51}{35.1} & \rgscore{-0.29}{47.4}/\rgscore{-0.53}{34.1} & \rgscore{-0.04}{61.3}/\rgscore{-0.08}{58.8} & \rgscore{-0.20}{52.3}/\rgscore{-0.41}{40.8} & \rgscore{-0.23}{50.6}/\rgscore{-0.42}{40.3} & \rgscore{-0.31}{46.0}/\rgscore{-0.34}{44.5} & \rgscore{-0.46}{37.9}/\rgscore{-0.66}{27.0} & \rgscore{-0.45}{38.4}/\rgscore{-0.64}{28.0} & \rgscore{-0.36}{43.4}/\rgscore{-0.53}{34.1} & \rgscore{-0.49}{36.2}/\rgscore{-0.71}{24.2} & \rgscore{-0.56}{32.3}/\rgscore{-0.69}{25.1} \\
 & Item & 46.5 & \rgscore{-0.10}{42.7}/\rgscore{-0.08}{43.6} & \rgscore{-0.23}{37.9}/\rgscore{-0.42}{30.8} & \rgscore{-0.33}{34.1}/\rgscore{-0.46}{29.4} & \rgscore{-0.13}{41.7}/\rgscore{-0.14}{41.2} & \rgscore{-0.15}{40.9}/\rgscore{-0.47}{28.9} & \rgscore{-0.26}{37.0}/\rgscore{-0.42}{30.8} & \rgscore{-0.44}{30.3}/\rgscore{-0.26}{37.0} & \rgscore{-0.55}{26.1}/\rgscore{-0.56}{25.6} & \rgscore{-0.49}{28.4}/\rgscore{-0.56}{25.6} & \rgscore{-0.41}{31.1}/\rgscore{-0.59}{24.6} & \rgscore{-0.52}{27.2}/\rgscore{-0.77}{18.0} & \rgscore{-0.56}{25.5}/\rgscore{-0.68}{21.3} \\
GPT-4.1 & Country & 65.0 & \rgscore{0.03}{66.8}/\rgscore{-0.14}{57.3} & \rgscore{-0.18}{55.0}/\rgscore{-0.59}{31.3} & \rgscore{-0.26}{50.2}/\rgscore{-0.62}{29.9} & \rgscore{-0.07}{60.9}/\rgscore{-0.16}{55.9} & \rgscore{-0.17}{55.3}/\rgscore{-0.35}{45.0} & \rgscore{-0.24}{51.5}/\rgscore{-0.40}{42.2} & \rgscore{-0.35}{45.0}/\rgscore{-0.49}{37.0} & \rgscore{-0.59}{31.3}/\rgscore{-0.80}{19.4} & \rgscore{-0.64}{28.4}/\rgscore{-0.77}{21.3} & \rgscore{-0.39}{43.0}/\rgscore{-0.59}{31.3} & \rgscore{-0.56}{33.2}/\rgscore{-0.75}{22.7} & \rgscore{-0.62}{29.8}/\rgscore{-0.78}{20.9} \\
 & Item & 57.2 & \rgscore{0.01}{57.8}/\rgscore{-0.07}{54.0} & \rgscore{-0.18}{48.8}/\rgscore{-0.36}{40.8} & \rgscore{-0.24}{46.0}/\rgscore{-0.38}{39.8} & \rgscore{-0.07}{54.0}/\rgscore{-0.15}{50.2} & \rgscore{-0.13}{51.1}/\rgscore{-0.37}{40.3} & \rgscore{-0.20}{48.1}/\rgscore{-0.43}{37.4} & \rgscore{-0.36}{40.8}/\rgscore{-0.40}{38.9} & \rgscore{-0.62}{28.9}/\rgscore{-0.60}{29.9} & \rgscore{-0.65}{27.5}/\rgscore{-0.63}{28.4} & \rgscore{-0.42}{37.9}/\rgscore{-0.64}{28.0} & \rgscore{-0.57}{31.1}/\rgscore{-0.79}{21.3} & \rgscore{-0.62}{28.9}/\rgscore{-0.85}{18.5} \\
Gemini-2.5-Pro & Country & 65.4 & \rgscore{0.00}{65.4}/\rgscore{-0.18}{55.0} & \rgscore{-0.03}{63.5}/\rgscore{-0.55}{34.1} & \rgscore{-0.09}{60.2}/\rgscore{-0.57}{32.7} & \rgscore{-0.06}{62.1}/\rgscore{-0.09}{60.2} & \rgscore{-0.10}{59.6}/\rgscore{-0.29}{48.8} & \rgscore{-0.13}{57.9}/\rgscore{-0.32}{47.4} & \rgscore{-0.16}{56.4}/\rgscore{-0.34}{46.0} & \rgscore{-0.38}{43.6}/\rgscore{-0.72}{24.2} & \rgscore{-0.33}{46.4}/\rgscore{-0.62}{29.9} & \rgscore{-0.24}{51.9}/\rgscore{-0.37}{44.5} & \rgscore{-0.39}{43.4}/\rgscore{-0.57}{32.7} & \rgscore{-0.33}{46.4}/\rgscore{-0.57}{32.7} \\
 & Item & 65.8 & \rgscore{-0.06}{62.6}/\rgscore{-0.06}{62.6} & \rgscore{-0.11}{60.2}/\rgscore{-0.35}{47.4} & \rgscore{-0.14}{58.3}/\rgscore{-0.36}{46.9} & \rgscore{-0.05}{63.4}/\rgscore{-0.19}{55.9} & \rgscore{-0.12}{59.6}/\rgscore{-0.30}{49.8} & \rgscore{-0.17}{56.6}/\rgscore{-0.38}{46.0} & \rgscore{-0.22}{54.0}/\rgscore{-0.22}{54.0} & \rgscore{-0.38}{46.0}/\rgscore{-0.46}{41.7} & \rgscore{-0.38}{46.0}/\rgscore{-0.38}{46.0} & \rgscore{-0.28}{51.1}/\rgscore{-0.40}{45.0} & \rgscore{-0.40}{44.7}/\rgscore{-0.60}{34.1} & \rgscore{-0.36}{46.8}/\rgscore{-0.58}{35.5} \\
Gemini-2.5-Flash & Country & 66.7 & \rgscore{-0.05}{64.0}/\rgscore{-0.25}{52.1} & \rgscore{-0.19}{55.9}/\rgscore{-0.63}{30.3} & \rgscore{-0.20}{55.0}/\rgscore{-0.67}{27.6} & \rgscore{-0.07}{62.8}/\rgscore{-0.14}{58.8} & \rgscore{-0.23}{53.4}/\rgscore{-0.40}{43.3} & \rgscore{-0.26}{51.5}/\rgscore{-0.37}{45.0} & \rgscore{-0.27}{51.2}/\rgscore{-0.38}{44.3} & \rgscore{-0.43}{41.7}/\rgscore{-0.79}{20.9} & \rgscore{-0.45}{40.3}/\rgscore{-0.72}{24.6} & \rgscore{-0.32}{48.1}/\rgscore{-0.45}{40.5} & \rgscore{-0.36}{46.0}/\rgscore{-0.63}{30.3} & \rgscore{-0.44}{41.0}/\rgscore{-0.59}{32.2} \\
 & Item & 64.2 & \rgscore{-0.09}{59.7}/\rgscore{-0.13}{57.3} & \rgscore{-0.11}{58.8}/\rgscore{-0.33}{47.4} & \rgscore{-0.15}{56.4}/\rgscore{-0.31}{48.1} & \rgscore{-0.10}{59.0}/\rgscore{-0.20}{54.0} & \rgscore{-0.19}{54.7}/\rgscore{-0.35}{46.2} & \rgscore{-0.24}{51.9}/\rgscore{-0.38}{44.5} & \rgscore{-0.31}{48.3}/\rgscore{-0.28}{50.0} & \rgscore{-0.39}{44.1}/\rgscore{-0.57}{35.1} & \rgscore{-0.42}{42.7}/\rgscore{-0.47}{40.3} & \rgscore{-0.29}{49.4}/\rgscore{-0.51}{38.1} & \rgscore{-0.36}{45.5}/\rgscore{-0.59}{34.1} & \rgscore{-0.42}{42.7}/\rgscore{-0.60}{33.6} \\
Claude-4.5-Sonnet & Country & 53.8 & \rgscore{-0.03}{52.6}/\rgscore{-0.04}{52.1} & \rgscore{-0.36}{37.1}/\rgscore{-0.48}{31.4} & \rgscore{-0.46}{32.4}/\rgscore{-0.52}{29.4} & \rgscore{-0.09}{49.6}/\rgscore{-0.04}{51.9} & \rgscore{-0.26}{41.6}/\rgscore{-0.50}{30.2} & \rgscore{-0.28}{40.5}/\rgscore{-0.53}{28.9} & \rgscore{-0.47}{31.9}/\rgscore{-0.45}{32.7} & \rgscore{-0.59}{26.1}/\rgscore{-0.63}{24.2} & \rgscore{-0.67}{22.3}/\rgscore{-0.65}{23.2} & \rgscore{-0.52}{29.4}/\rgscore{-0.69}{21.4} & \rgscore{-0.67}{22.1}/\rgscore{-0.77}{17.5} & \rgscore{-0.62}{24.9}/\rgscore{-0.83}{14.8} \\
 & Item & 49.6 & \rgscore{-0.03}{48.3}/\rgscore{-0.01}{49.3} & \rgscore{-0.41}{33.3}/\rgscore{-0.39}{34.3} & \rgscore{-0.47}{31.0}/\rgscore{-0.44}{32.2} & \rgscore{-0.10}{45.7}/\rgscore{-0.10}{45.7} & \rgscore{-0.29}{38.2}/\rgscore{-0.52}{28.8} & \rgscore{-0.30}{37.9}/\rgscore{-0.57}{27.0} & \rgscore{-0.51}{29.5}/\rgscore{-0.36}{35.5} & \rgscore{-0.62}{25.1}/\rgscore{-0.54}{28.0} & \rgscore{-0.72}{20.9}/\rgscore{-0.63}{24.6} & \rgscore{-0.54}{28.1}/\rgscore{-0.72}{21.0} & \rgscore{-0.72}{21.2}/\rgscore{-0.78}{18.5} & \rgscore{-0.66}{23.6}/\rgscore{-0.87}{15.3} \\
\midrule
\textbf{Open-Source} \\
\midrule
Qwen3-VL-30B & Country & 52.3 & \rgscore{0.01}{52.6}/\rgscore{-0.11}{47.4} & \rgscore{-0.39}{34.6}/\rgscore{-0.72}{19.4} & \rgscore{-0.62}{24.2}/\rgscore{-0.78}{16.6} & \rgscore{-0.05}{49.8}/\rgscore{-0.11}{47.4} & \rgscore{-0.33}{37.4}/\rgscore{-0.53}{28.0} & \rgscore{-0.46}{31.5}/\rgscore{-0.74}{18.5} & \rgscore{-0.30}{38.4}/\rgscore{-0.46}{31.3} & \rgscore{-0.57}{26.5}/\rgscore{-0.82}{14.7} & \rgscore{-0.66}{22.3}/\rgscore{-0.81}{15.2} & \rgscore{-0.40}{34.0}/\rgscore{-0.54}{27.5} & \rgscore{-0.54}{27.7}/\rgscore{-0.77}{17.1} & \rgscore{-0.66}{22.1}/\rgscore{-0.90}{11.4} \\
 & Item & 42.0 & \rgscore{0.01}{42.2}/\rgscore{-0.11}{38.4} & \rgscore{-0.26}{33.2}/\rgscore{-0.47}{26.1} & \rgscore{-0.41}{28.4}/\rgscore{-0.43}{27.5} & \rgscore{-0.06}{40.0}/\rgscore{-0.14}{37.4} & \rgscore{-0.29}{32.3}/\rgscore{-0.52}{24.6} & \rgscore{-0.35}{30.2}/\rgscore{-0.57}{22.7} & \rgscore{-0.39}{28.9}/\rgscore{-0.29}{32.2} & \rgscore{-0.55}{23.7}/\rgscore{-0.56}{23.2} & \rgscore{-0.64}{20.4}/\rgscore{-0.64}{20.4} & \rgscore{-0.44}{27.2}/\rgscore{-0.53}{24.2} & \rgscore{-0.50}{25.1}/\rgscore{-0.71}{18.0} & \rgscore{-0.77}{16.2}/\rgscore{-0.87}{12.8} \\
Qwen3-VL-4B & Country & 49.8 & \rgscore{-0.01}{49.3}/\rgscore{-0.07}{46.9} & \rgscore{-0.56}{25.6}/\rgscore{-0.72}{18.5} & \rgscore{-0.58}{24.6}/\rgscore{-0.80}{15.2} & \rgscore{-0.09}{46.0}/\rgscore{-0.10}{45.5} & \rgscore{-0.39}{32.8}/\rgscore{-0.60}{23.7} & \rgscore{-0.52}{27.2}/\rgscore{-0.66}{21.3} & \rgscore{-0.29}{37.0}/\rgscore{-0.29}{37.4} & \rgscore{-0.72}{18.5}/\rgscore{-0.83}{13.7} & \rgscore{-0.60}{23.7}/\rgscore{-0.74}{17.5} & \rgscore{-0.39}{32.8}/\rgscore{-0.46}{29.9} & \rgscore{-0.72}{18.3}/\rgscore{-0.76}{16.6} & \rgscore{-0.63}{22.6}/\rgscore{-0.81}{14.7} \\
 & Item & 32.9 & \rgscore{-0.10}{30.3}/\rgscore{0.01}{33.2} & \rgscore{-0.33}{24.2}/\rgscore{-0.32}{24.6} & \rgscore{-0.35}{23.7}/\rgscore{-0.35}{23.7} & \rgscore{-0.07}{31.1}/\rgscore{-0.21}{27.5} & \rgscore{-0.26}{26.0}/\rgscore{-0.48}{20.4} & \rgscore{-0.39}{22.6}/\rgscore{-0.49}{19.9} & \rgscore{-0.22}{27.0}/\rgscore{-0.26}{26.1} & \rgscore{-0.46}{20.9}/\rgscore{-0.57}{18.0} & \rgscore{-0.44}{21.3}/\rgscore{-0.49}{19.9} & \rgscore{-0.35}{23.8}/\rgscore{-0.44}{21.3} & \rgscore{-0.62}{16.6}/\rgscore{-0.62}{16.6} & \rgscore{-0.8}{16.6}/\rgscore{-0.75}{13.3} \\
Qwen3-VL-8B & Country & 44.9 & \rgscore{-0.08}{41.7}/\rgscore{-0.17}{38.4} & \rgscore{-0.88}{10.4}/\rgscore{-0.94}{8.1} & \rgscore{-0.90}{9.5}/\rgscore{-0.94}{8.1} & \rgscore{-0.14}{39.6}/\rgscore{-0.17}{38.4} & \rgscore{-0.77}{14.9}/\rgscore{-0.89}{10.0} & \rgscore{-0.78}{14.5}/\rgscore{-0.92}{9.0} & \rgscore{-0.36}{30.8}/\rgscore{-0.48}{26.1} & \rgscore{-0.92}{9.0}/\rgscore{-1.00}{5.7} & \rgscore{-0.83}{12.3}/\rgscore{-0.94}{8.1} & \rgscore{-0.56}{23.0}/\rgscore{-0.60}{21.3} & \rgscore{-0.88}{10.6}/\rgscore{-0.94}{8.1} & \rgscore{-0.83}{12.3}/\rgscore{-0.98}{6.6} \\
 & Item & 32.9 & \rgscore{-0.13}{29.4}/\rgscore{-0.15}{28.9} & \rgscore{-0.46}{20.9}/\rgscore{-0.49}{19.9} & \rgscore{-0.44}{21.3}/\rgscore{-0.42}{21.8} & \rgscore{-0.13}{29.4}/\rgscore{-0.17}{28.4} & \rgscore{-0.49}{20.0}/\rgscore{-0.57}{18.0} & \rgscore{-0.41}{22.1}/\rgscore{-0.64}{16.1} & \rgscore{-0.42}{21.8}/\rgscore{-0.40}{22.3} & \rgscore{-0.80}{11.8}/\rgscore{-0.75}{13.3} & \rgscore{-0.73}{13.7}/\rgscore{-0.82}{11.4} & \rgscore{-0.62}{16.6}/\rgscore{-0.71}{14.2} & \rgscore{-0.76}{12.8}/\rgscore{-0.87}{10.0} & \rgscore{-0.83}{11.1}/\rgscore{-1.00}{6.6} \\
InternVL3.5-38B & Country & 25.5 & \rgscore{0.04}{26.5}/\rgscore{-0.08}{23.7} & \rgscore{-0.63}{11.4}/\rgscore{-0.80}{7.6} & \rgscore{-0.66}{10.9}/\rgscore{-0.87}{6.2} & \rgscore{-0.17}{21.7}/\rgscore{-0.14}{22.3} & \rgscore{-0.57}{12.8}/\rgscore{-0.74}{9.0} & \rgscore{-0.63}{11.5}/\rgscore{-0.83}{7.1} & \rgscore{-0.38}{17.1}/\rgscore{-0.53}{13.7} & \rgscore{-0.78}{8.1}/\rgscore{-0.83}{7.1} & \rgscore{-0.72}{9.5}/\rgscore{-0.87}{6.2} & \rgscore{-0.48}{14.9}/\rgscore{-0.53}{13.7} & \rgscore{-0.76}{8.5}/\rgscore{-0.89}{5.7} & \rgscore{-0.82}{7.2}/\rgscore{-0.85}{6.6} \\
 & Item & 19.8 & \rgscore{0.09}{21.3}/\rgscore{0.07}{20.9} & \rgscore{-0.41}{13.3}/\rgscore{-0.47}{12.3} & \rgscore{-0.53}{11.4}/\rgscore{-0.51}{11.8} & \rgscore{-0.37}{14.0}/\rgscore{-0.11}{18.0} & \rgscore{-0.50}{11.9}/\rgscore{-0.62}{10.0} & \rgscore{-0.63}{9.8}/\rgscore{-0.74}{8.1} & \rgscore{-0.32}{14.7}/\rgscore{-0.56}{10.9} & \rgscore{-0.74}{8.1}/\rgscore{-0.68}{9.0} & \rgscore{-0.71}{8.5}/\rgscore{-0.68}{9.0} & \rgscore{-0.55}{11.1}/\rgscore{-0.56}{10.9} & \rgscore{-0.80}{7.2}/\rgscore{-0.86}{6.2} & \rgscore{-0.98}{4.3}/\rgscore{-0.89}{5.7} \\
InternVL3.5-4B & Country & 20.2 & \rgscore{-0.15}{17.5}/\rgscore{-0.07}{19.0} & \rgscore{-0.61}{9.5}/\rgscore{-0.77}{6.6} & \rgscore{-0.82}{5.7}/\rgscore{-0.82}{5.7} & \rgscore{-0.16}{17.4}/\rgscore{-0.18}{17.1} & \rgscore{-0.54}{10.6}/\rgscore{-0.66}{8.5} & \rgscore{-0.66}{8.5}/\rgscore{-0.79}{6.2} & \rgscore{-0.53}{10.9}/\rgscore{-0.34}{14.2} & \rgscore{-0.85}{5.2}/\rgscore{-0.85}{5.2} & \rgscore{-0.88}{4.7}/\rgscore{-0.93}{3.8} & \rgscore{-0.61}{9.4}/\rgscore{-0.56}{10.4} & \rgscore{-0.76}{6.8}/\rgscore{-0.85}{5.2} & \rgscore{-0.88}{4.7}/\rgscore{-0.93}{3.8} \\
 & Item & 11.9 & \rgscore{-0.16}{10.4}/\rgscore{-0.05}{11.4} & \rgscore{-0.65}{5.7}/\rgscore{-0.60}{6.2} & \rgscore{-0.85}{3.8}/\rgscore{-0.50}{7.1} & \rgscore{0.00}{11.9}/\rgscore{-0.30}{9.0} & \rgscore{-0.58}{6.4}/\rgscore{-0.76}{4.7} & \rgscore{-0.67}{5.5}/\rgscore{-0.85}{3.8} & \rgscore{-0.50}{7.1}/\rgscore{-0.45}{7.6} & \rgscore{-0.76}{4.7}/\rgscore{-0.45}{7.6} & \rgscore{-0.70}{5.2}/\rgscore{-0.76}{4.7} & \rgscore{-0.85}{3.8}/\rgscore{-0.70}{5.2} & \rgscore{-0.58}{6.4}/\rgscore{-0.80}{4.3} & \rgscore{-0.89}{3.4}/\rgscore{-0.96}{2.8} \\
InternVL3.5-8B & Country & 21.0 & \rgscore{0.12}{23.2}/\rgscore{0.04}{21.8} & \rgscore{-0.68}{8.5}/\rgscore{-0.73}{7.6} & \rgscore{-0.76}{7.1}/\rgscore{-0.83}{5.7} & \rgscore{-0.13}{18.7}/\rgscore{-0.09}{19.4} & \rgscore{-0.80}{6.4}/\rgscore{-0.60}{10.0} & \rgscore{-0.91}{4.3}/\rgscore{-0.68}{8.5} & \rgscore{-0.50}{11.8}/\rgscore{-0.47}{12.3} & \rgscore{-0.83}{5.7}/\rgscore{-0.83}{5.7} & \rgscore{-0.91}{4.3}/\rgscore{-0.76}{7.1} & \rgscore{-0.59}{10.2}/\rgscore{-0.70}{8.1} & \rgscore{-0.80}{6.4}/\rgscore{-0.89}{4.7} & \rgscore{-0.91}{4.3}/\rgscore{-0.83}{5.7} \\
 & Item & 12.3 & \rgscore{0.24}{14.7}/\rgscore{0.19}{14.2} & \rgscore{-0.72}{5.2}/\rgscore{-0.62}{6.2} & \rgscore{-0.62}{6.2}/\rgscore{-0.67}{5.7} & \rgscore{0.00}{12.3}/\rgscore{-0.05}{11.8} & \rgscore{-0.69}{5.5}/\rgscore{-0.58}{6.6} & \rgscore{-0.77}{4.7}/\rgscore{-0.58}{6.6} & \rgscore{-0.67}{5.7}/\rgscore{-0.34}{9.0} & \rgscore{-0.92}{3.3}/\rgscore{-0.77}{4.7} & \rgscore{-0.86}{3.8}/\rgscore{-0.62}{6.2} & \rgscore{-0.77}{4.7}/\rgscore{-0.67}{5.7} & \rgscore{-0.81}{4.3}/\rgscore{-0.92}{3.3} & \rgscore{-0.86}{3.8}/\rgscore{-0.81}{4.3} \\
\bottomrule
\end{tabular}%
}
\end{table*}

\begin{table*}[!t]
\centering
\caption{Distraction rate (\%) across models under different perturbation settings. Columns indicate the percentage of wrong predictions that were distracted by the adversarial country across landmark (L), flag (F), and both (L+F) perturbation context infusion.}
\label{tab:complete-distraction}
\resizebox{\textwidth}{!}{
\begin{tabular}{lcccccccccccc}
\toprule
\multirow{3}{*}{Model} & \multicolumn{6}{c}{Naive} & \multicolumn{6}{c}{AI} \\
\cmidrule(lr){2-7} \cmidrule(lr){8-13}
& \multicolumn{3}{c}{Geo} & \multicolumn{3}{c}{Desc}
& \multicolumn{3}{c}{Geo} & \multicolumn{3}{c}{Desc} \\
\cmidrule(lr){2-4} \cmidrule(lr){5-7}
\cmidrule(lr){8-10} \cmidrule(lr){11-13}
& L & F & L+F & L & F & L+F & L & F & L+F & L & F & L+F \\
\midrule
\textit{Difficulty} & Easy/Hard & Easy/Hard & Easy/Hard & Easy/Hard & Easy/Hard & Easy/Hard & Easy/Hard & Easy/Hard & Easy/Hard & Easy/Hard & Easy/Hard & Easy/Hard \\
\midrule
\textbf{Proprietary} \\
\midrule
GPT-5 (High) & \bluehl{8.5}{8.5}/\bluehl{37.2}{37.2} & \bluehl{33.3}{33.3}/\bluehl{75.5}{75.5} & \bluehl{29.7}{29.7}/\bluehl{78.1}{78.1} & \bluehl{10.1}{10.1}/\bluehl{36.0}{36.0} & \bluehl{23.7}{23.7}/\bluehl{60.0}{60.0} & \bluehl{25.3}{25.3}/\bluehl{60.8}{60.8} & \bluehl{16.2}{16.2}/\bluehl{48.1}{48.1} & \bluehl{41.1}{41.1}/\bluehl{72.5}{72.5} & \bluehl{35.1}{35.1}/\bluehl{75.6}{75.6} & \bluehl{21.3}{21.3}/\bluehl{43.7}{43.7} & \bluehl{36.1}{36.1}/\bluehl{71.9}{71.9} & \bluehl{33.1}{33.1}/\bluehl{65.7}{65.7} \\
GPT-5 (Low) & \bluehl{8.4}{8.4}/\bluehl{26.7}{26.7} & \bluehl{34.8}{34.8}/\bluehl{66.7}{66.7} & \bluehl{30.3}{30.3}/\bluehl{67.3}{67.3} & \bluehl{11.8}{11.8}/\bluehl{32.6}{32.6} & \bluehl{20.4}{20.4}/\bluehl{58.9}{58.9} & \bluehl{22.3}{22.3}/\bluehl{62.3}{62.3} & \bluehl{22.1}{22.1}/\bluehl{46.8}{46.8} & \bluehl{41.3}{41.3}/\bluehl{69.5}{69.5} & \bluehl{31.7}{31.7}/\bluehl{70.2}{70.2} & \bluehl{21.8}{21.8}/\bluehl{41.5}{41.5} & \bluehl{40.1}{40.1}/\bluehl{72.0}{72.0} & \bluehl{38.5}{38.5}/\bluehl{65.5}{65.5} \\
GPT-5 (Minimal) & \bluehl{10.3}{10.3}/\bluehl{24.7}{24.7} & \bluehl{57.8}{57.8}/\bluehl{80.3}{80.3} & \bluehl{54.1}{54.1}/\bluehl{82.7}{82.7} & \bluehl{9.9}{9.9}/\bluehl{33.3}{33.3} & \bluehl{43.8}{43.8}/\bluehl{75.2}{75.2} & \bluehl{50.0}{50.0}/\bluehl{73.8}{73.8} & \bluehl{36.8}{36.8}/\bluehl{58.1}{58.1} & \bluehl{64.1}{64.1}/\bluehl{82.5}{82.5} & \bluehl{63.8}{63.8}/\bluehl{86.8}{86.8} & \bluehl{36.8}{36.8}/\bluehl{56.1}{56.1} & \bluehl{62.7}{62.7}/\bluehl{80.6}{80.6} & \bluehl{50.3}{50.3}/\bluehl{77.8}{77.8} \\
GPT-4.1 & \bluehl{15.7}{15.7}/\bluehl{45.6}{45.6} & \bluehl{60.0}{60.0}/\bluehl{91.7}{91.7} & \bluehl{61.0}{61.0}/\bluehl{93.2}{93.2} & \bluehl{17.4}{17.4}/\bluehl{41.9}{41.9} & \bluehl{47.6}{47.6}/\bluehl{80.2}{80.2} & \bluehl{49.1}{49.1}/\bluehl{82.8}{82.8} & \bluehl{35.3}{35.3}/\bluehl{75.9}{75.9} & \bluehl{77.2}{77.2}/\bluehl{94.7}{94.7} & \bluehl{76.2}{76.2}/\bluehl{97.0}{97.0} & \bluehl{46.3}{46.3}/\bluehl{69.7}{69.7} & \bluehl{73.9}{73.9}/\bluehl{89.6}{89.6} & \bluehl{66.7}{66.7}/\bluehl{89.2}{89.2} \\
Gemini-2.5-Pro & \bluehl{18.4}{18.4}/\bluehl{55.4}{55.4} & \bluehl{57.0}{57.0}/\bluehl{81.6}{81.6} & \bluehl{65.3}{65.3}/\bluehl{82.2}{82.2} & \bluehl{20.7}{20.7}/\bluehl{49.4}{49.4} & \bluehl{52.8}{52.8}/\bluehl{74.8}{74.8} & \bluehl{54.4}{54.4}/\bluehl{78.4}{78.4} & \bluehl{37.9}{37.9}/\bluehl{69.2}{69.2} & \bluehl{68.3}{68.3}/\bluehl{85.6}{85.6} & \bluehl{69.0}{69.0}/\bluehl{84.3}{84.3} & \bluehl{34.7}{34.7}/\bluehl{60.0}{60.0} & \bluehl{66.1}{66.1}/\bluehl{78.2}{78.2} & \bluehl{52.2}{52.2}/\bluehl{81.8}{81.8} \\
Claude-4.5-Sonnet & \bluehl{9.0}{9.0}/\bluehl{35.6}{35.6} & \bluehl{67.4}{67.4}/\bluehl{85.4}{85.4} & \bluehl{65.5}{65.5}/\bluehl{86.6}{86.6} & \bluehl{10.3}{10.3}/\bluehl{34.0}{34.0} & \bluehl{55.1}{55.1}/\bluehl{75.5}{75.5} & \bluehl{55.1}{55.1}/\bluehl{80.0}{80.0} & \bluehl{39.9}{39.9}/\bluehl{52.8}{52.8} & \bluehl{65.4}{65.4}/\bluehl{85.0}{85.0} & \bluehl{57.9}{57.9}/\bluehl{79.0}{79.0} & \bluehl{40.4}{40.4}/\bluehl{56.4}{56.4} & \bluehl{68.9}{68.9}/\bluehl{79.9}{79.9} & \bluehl{59.4}{59.4}/\bluehl{77.0}{77.0} \\
\midrule
\textbf{Open-Source} \\
\midrule
Qwen3-VL-30B & \bluehl{12.0}{12.0}/\bluehl{46.8}{46.8} & \bluehl{85.5}{85.5}/\bluehl{95.9}{95.9} & \bluehl{90.0}{90.0}/\bluehl{98.9}{98.9} & \bluehl{18.6}{18.6}/\bluehl{33.3}{33.3} & \bluehl{76.2}{76.2}/\bluehl{90.1}{90.1} & \bluehl{88.8}{88.8}/\bluehl{95.3}{95.3} & \bluehl{50.0}{50.0}/\bluehl{73.1}{73.1} & \bluehl{85.8}{85.8}/\bluehl{97.2}{97.2} & \bluehl{89.6}{89.6}/\bluehl{97.8}{97.8} & \bluehl{45.2}{45.2}/\bluehl{63.4}{63.4} & \bluehl{79.4}{79.4}/\bluehl{90.9}{90.9} & \bluehl{79.2}{79.2}/\bluehl{93.0}{93.0} \\
Qwen3-VL-4B & \bluehl{14.0}{14.0}/\bluehl{38.4}{38.4} & \bluehl{90.4}{90.4}/\bluehl{91.9}{91.9} & \bluehl{90.6}{90.6}/\bluehl{91.6}{91.6} & \bluehl{15.0}{15.0}/\bluehl{35.7}{35.7} & \bluehl{86.7}{86.7}/\bluehl{88.8}{88.8} & \bluehl{86.5}{86.5}/\bluehl{89.8}{89.8} & \bluehl{42.9}{42.9}/\bluehl{54.5}{54.5} & \bluehl{90.7}{90.7}/\bluehl{94.0}{94.0} & \bluehl{75.8}{75.8}/\bluehl{92.0}{92.0} & \bluehl{45.6}{45.6}/\bluehl{52.7}{52.7} & \bluehl{89.6}{89.6}/\bluehl{89.8}{89.8} & \bluehl{84.1}{84.1}/\bluehl{85.6}{85.6} \\
Qwen3-VL-8B & \bluehl{26.8}{26.8}/\bluehl{50.8}{50.8} & \bluehl{94.2}{94.2}/\bluehl{94.3}{94.3} & \bluehl{94.2}{94.2}/\bluehl{96.4}{96.4} & \bluehl{26.8}{26.8}/\bluehl{43.1}{43.1} & \bluehl{94.5}{94.5}/\bluehl{94.7}{94.7} & \bluehl{96.0}{96.0}/\bluehl{95.3}{95.3} & \bluehl{58.2}{58.2}/\bluehl{67.9}{67.9} & \bluehl{95.3}{95.3}/\bluehl{97.0}{97.0} & \bluehl{93.5}{93.5}/\bluehl{97.4}{97.4} & \bluehl{63.5}{63.5}/\bluehl{62.0}{62.0} & \bluehl{96.2}{96.2}/\bluehl{94.8}{94.8} & \bluehl{94.7}{94.7}/\bluehl{93.4}{93.4} \\
InternVL3.5-38B & \bluehl{12.9}{12.9}/\bluehl{37.9}{37.9} & \bluehl{47.6}{47.6}/\bluehl{63.1}{63.1} & \bluehl{59.6}{59.6}/\bluehl{74.7}{74.7} & \bluehl{23.4}{23.4}/\bluehl{28.0}{28.0} & \bluehl{72.2}{72.2}/\bluehl{58.9}{58.9} & \bluehl{79.8}{79.8}/\bluehl{66.8}{66.8} & \bluehl{36.6}{36.6}/\bluehl{46.7}{46.7} & \bluehl{56.2}{56.2}/\bluehl{70.4}{70.4} & \bluehl{76.4}{76.4}/\bluehl{72.7}{72.7} & \bluehl{41.5}{41.5}/\bluehl{51.1}{51.1} & \bluehl{80.0}{80.0}/\bluehl{62.8}{62.8} & \bluehl{82.6}{82.6}/\bluehl{72.6}{72.6} \\
InternVL3.5-4B & \bluehl{9.8}{9.8}/\bluehl{31.0}{31.0} & \bluehl{42.4}{42.4}/\bluehl{59.4}{59.4} & \bluehl{46.7}{46.7}/\bluehl{61.3}{61.3} & \bluehl{27.8}{27.8}/\bluehl{24.0}{24.0} & \bluehl{75.2}{75.2}/\bluehl{56.5}{56.5} & \bluehl{77.7}{77.7}/\bluehl{59.6}{59.6} & \bluehl{27.1}{27.1}/\bluehl{45.9}{45.9} & \bluehl{43.0}{43.0}/\bluehl{56.5}{56.5} & \bluehl{50.7}{50.7}/\bluehl{67.0}{67.0} & \bluehl{47.9}{47.9}/\bluehl{42.3}{42.3} & \bluehl{76.7}{76.7}/\bluehl{59.0}{59.0} & \bluehl{82.6}{82.6}/\bluehl{67.5}{67.5} \\
InternVL3.5-8B & \bluehl{6.8}{6.8}/\bluehl{26.1}{26.1} & \bluehl{50.3}{50.3}/\bluehl{56.4}{56.4} & \bluehl{55.6}{55.6}/\bluehl{63.8}{63.8} & \bluehl{19.4}{19.4}/\bluehl{24.1}{24.1} & \bluehl{80.5}{80.5}/\bluehl{62.1}{62.1} & \bluehl{82.7}{82.7}/\bluehl{66.3}{66.3} & \bluehl{25.8}{25.8}/\bluehl{42.2}{42.2} & \bluehl{44.7}{44.7}/\bluehl{64.8}{64.8} & \bluehl{62.9}{62.9}/\bluehl{76.0}{76.0} & \bluehl{44.1}{44.1}/\bluehl{44.3}{44.3} & \bluehl{75.9}{75.9}/\bluehl{59.7}{59.7} & \bluehl{82.7}{82.7}/\bluehl{75.4}{75.4} \\
\bottomrule
\end{tabular}%
}
\end{table*}

We present the complete results in Table~\ref{tab:complete-accuracy}. The key observations are consistent with the findings reported in the main paper, demonstrating that the conclusions generalize across all evaluated settings.

\begin{itemize}
    \item \textbf{Proprietary VLMs outperform open-source models}, with the exception of Claude, which performs comparably to Qwen3-VL.
    \item \textbf{Flag perturbation} emerges as the dominant perturbation factor, exhibiting a stronger influence than landmark perturbations.
    \item \textbf{Generative perturbation} consistently outperforms image stacking across all cases, highlighting its effectiveness as a perturbation strategy.
    \item \textbf{Geographic (Geo) and Descriptive (Desc) proxies show similar patterns in all cases}, suggesting that both proxy-based semantic and geographic cultural cues remain relevant.
\end{itemize}

\section{Inference Model Selection \& Hyperparmeters}
\label{appendix:model-hyperparams}
All models are evaluated with a temperature of 0.0 to ensure deterministic outputs.
Structured output formatting is used throughout. For GPT and Gemini, structured JSON responses are enabled directly through their respective API features. For other models, we explicitly instruct the model to produce valid JSON outputs via prompt formatting.
The open-source model checkpoints used are as follows:
\begin{itemize}
\item \texttt{OpenGVLab/InternVL3\_5-38B}
\item \texttt{OpenGVLab/InternVL3\_5-8B}
\item \texttt{OpenGVLab/InternVL3\_5-4B}
\item \texttt{Qwen/Qwen3-VL-30B-A3B-Instruct}
\item \texttt{Qwen/Qwen3-VL-8B-A3B-Instruct}
\item \texttt{Qwen/Qwen3-VL-4B-A3B-Instruct}
\end{itemize}

\noindent For the proprietary models, we employed different settings for each variant to assess their impact on performance and robustness. As with the open-weight variants, the \textit{sampling temperature is uniformly set to $0$} across all proprietary models to ensure deterministic output.

\begin{itemize}
    \item \texttt{GPT-5-H+} (GPT-5): In this setting, we utilize \textit{high-level thinking} and \textit{high verbosity}. This configuration serves as the \textit{upper-bound variant} to test the maximum reasoning capacity and its effect on the model's robustness toward the \confusedtourist scenario.
    \item \texttt{GPT-5-H} (GPT-5): This configuration uses \textit{high-level thinking} but with a \textit{standard verbosity} setting.
    \item \texttt{GPT-5-L} (GPT-5): This configuration utilizes \textit{low-level thinking} with \textit{standard verbosity}.
    \item \texttt{GPT-5-m} (GPT-5): This setting uses \textit{minimum-level thinking} and \textit{standard verbosity}.
    \item \texttt{GPT-4.1} (GPT-4.1): This model is configured with \textit{standard verbosity} and \textit{standard-level thinking}.
    \item \texttt{G-2.5-Pro} (Gemini 2.5 Pro): For this variant, the model's \textit{dynamic thinking budget is set to $-1$} (denoting the highest available thinking budget) with \textit{standard verbosity}.
    \item \texttt{G-2.5-flash} (Gemini 2.5 Flash): This model is configured to use \textit{no internal thinking at all}, with the \textit{dynamic thinking budget set to $0$}, and operates with \textit{standard verbosity}.
    \item \texttt{Sonnet-4.5} (Claude 4.5 Sonnet): This variant is tested using its \textit{default configuration settings} as provided by the vendor.
\end{itemize}

All models are evaluated under identical inference settings, unless otherwise specified.

\section{Prompt Ablation Attempt}
\label{appendix:token-ablation-study}

To mitigate hallucination and misgrounding issues observed during evaluation, we conducted a \textit{prompt ablation study} aimed at reducing over-attention to perturbational cues. The core idea is to remove prompt tokens that may induce attention overload toward contextual or background signals, which can cause vision-language models (VLMs) to rely excessively on spurious visual elements (e.g., flags or landmarks). Instead, we introduced a more explicit focusing instruction directing the model to concentrate on the target object itself:
\texttt{"Identify the traditional name and origin of the {category} in the image. Please identify based on the object itself, not from surrounding cues."}

\begin{figure*}
\centering
\includegraphics[width=\linewidth]{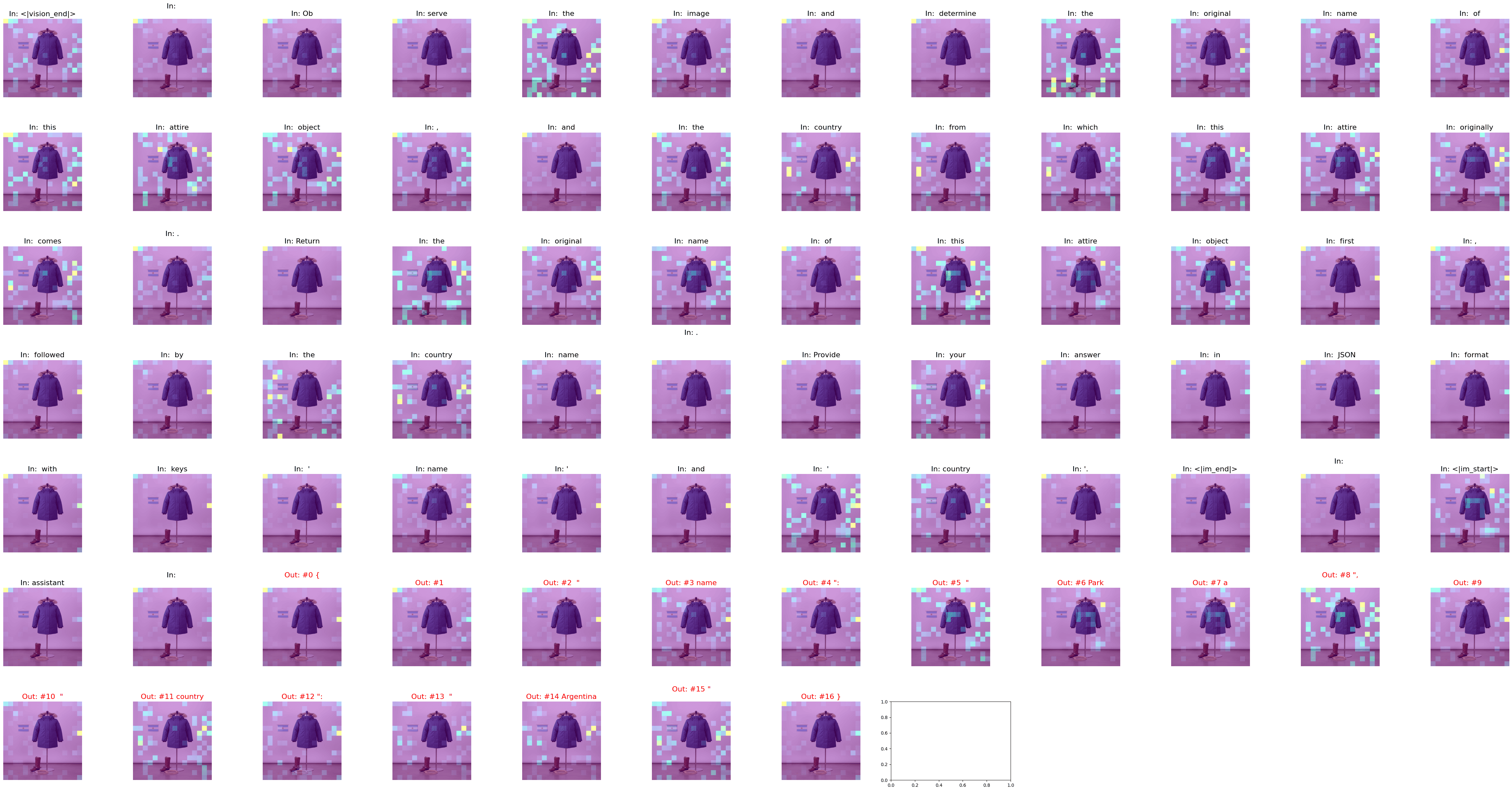}
\caption{Original attire prompt, which has right country (\textbf{wrong}).}
\label{fig:ori_attire}
\end{figure*}

\begin{figure*}
\centering
\includegraphics[width=\linewidth]{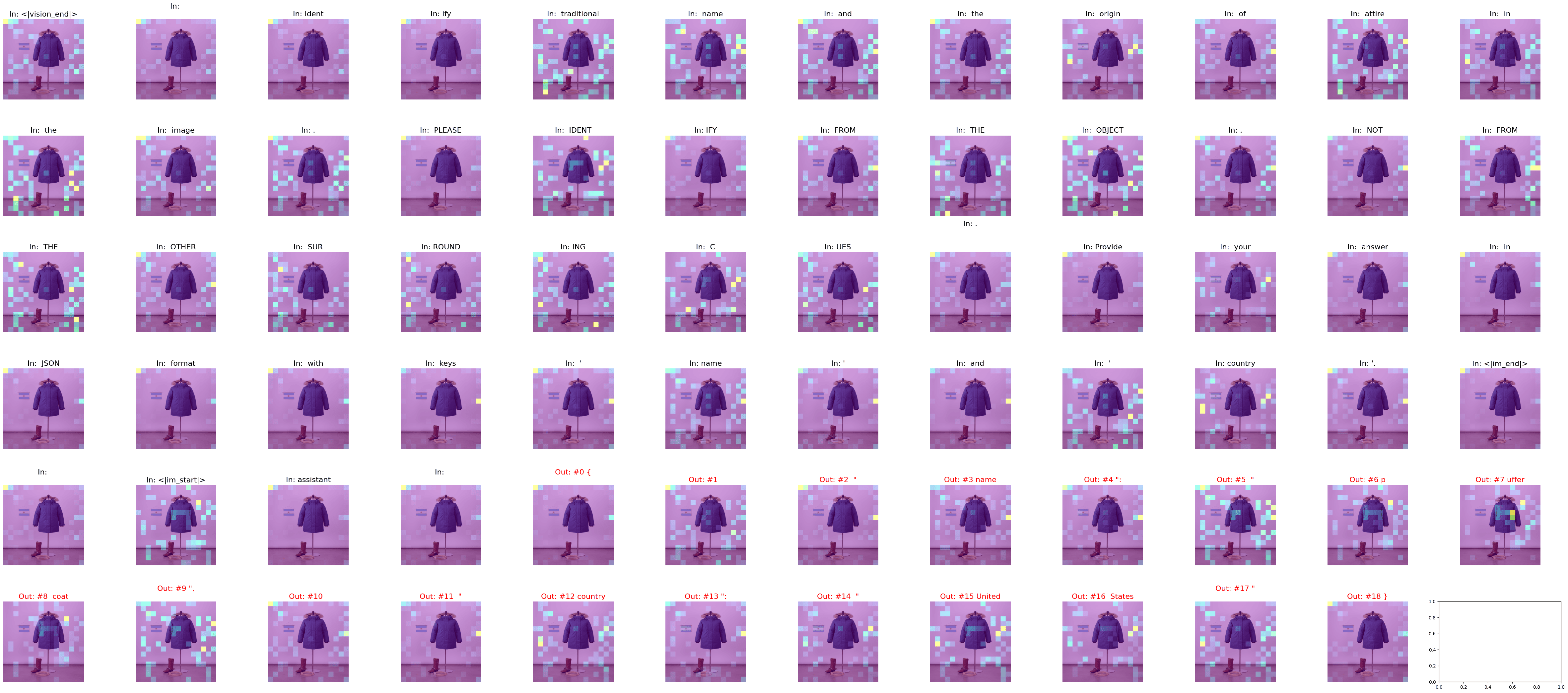}
\caption{Refined attire prompt, this is instability case where now, instead of just country, both name and country is wrong.  (\textbf{wrong}).}
\label{fig:fixed_attire}
\end{figure*}

\begin{figure*}
\centering
\includegraphics[width=\linewidth]{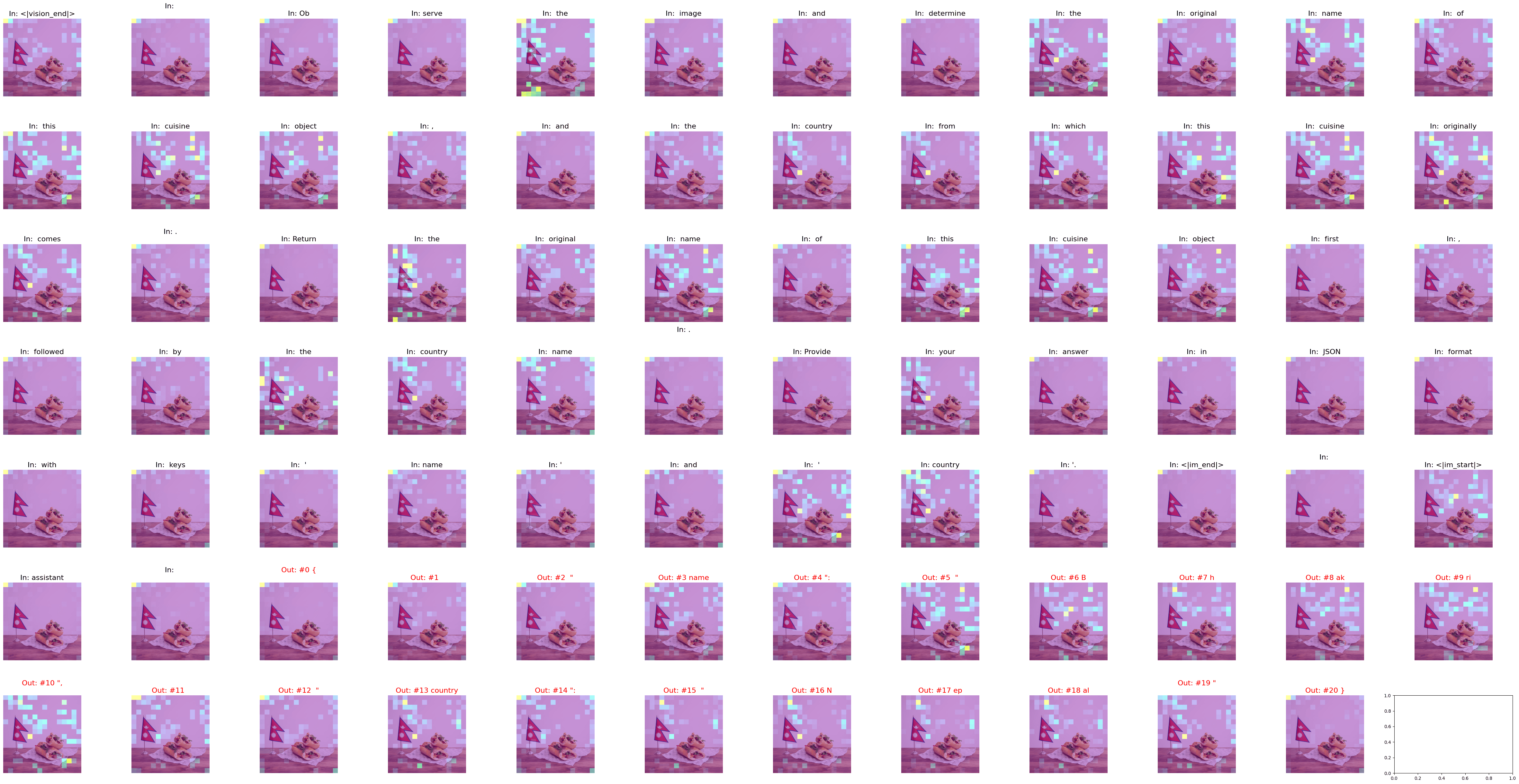}
\caption{Original cuisine prompt (\textbf{wrong}).}
\label{fig:ori_cuisine}
\end{figure*}

\begin{figure*}
\centering
\includegraphics[width=\linewidth]{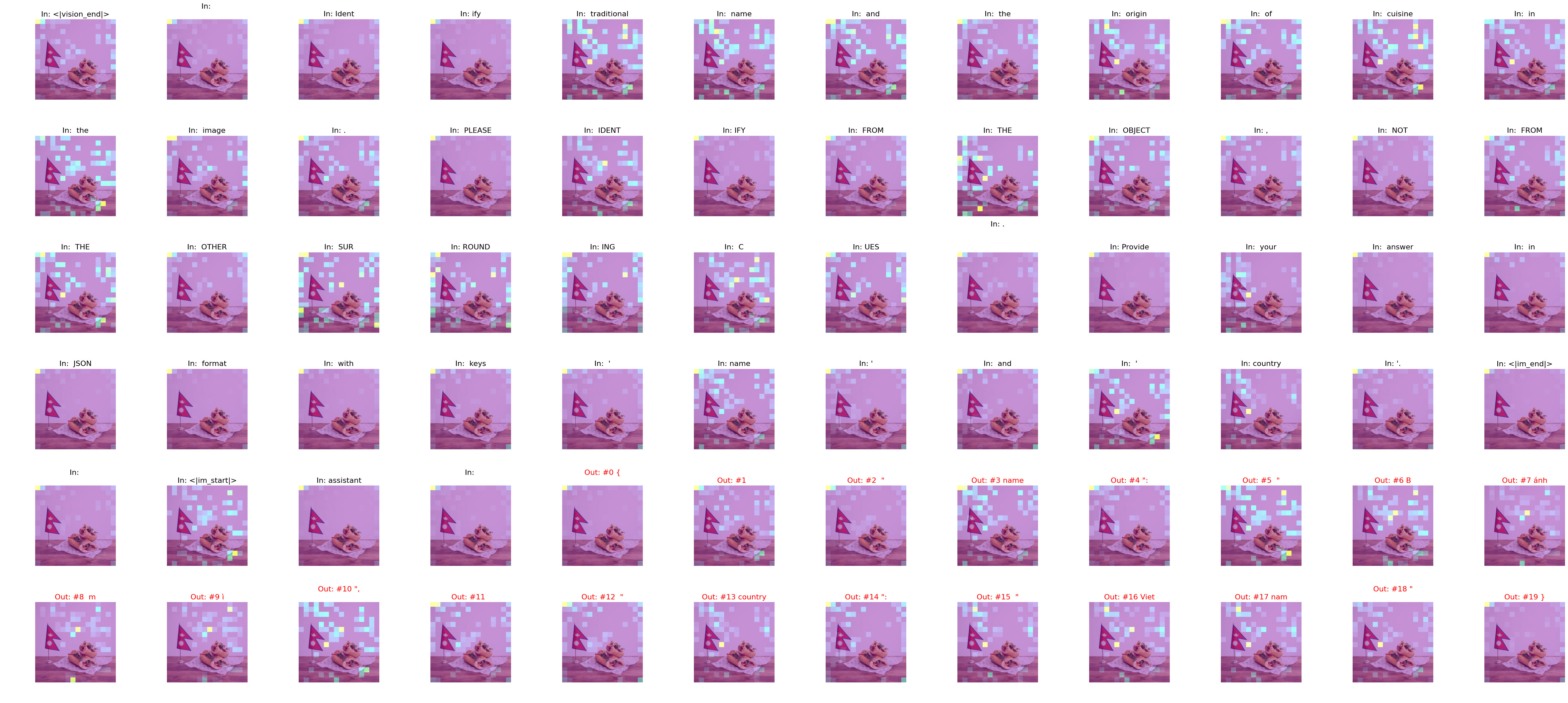}
\caption{Refined cuisine prompt (\textbf{correct}).}
\label{fig:fixed_cuisine}
\end{figure*}

\begin{figure*}
\centering
\includegraphics[width=\linewidth]{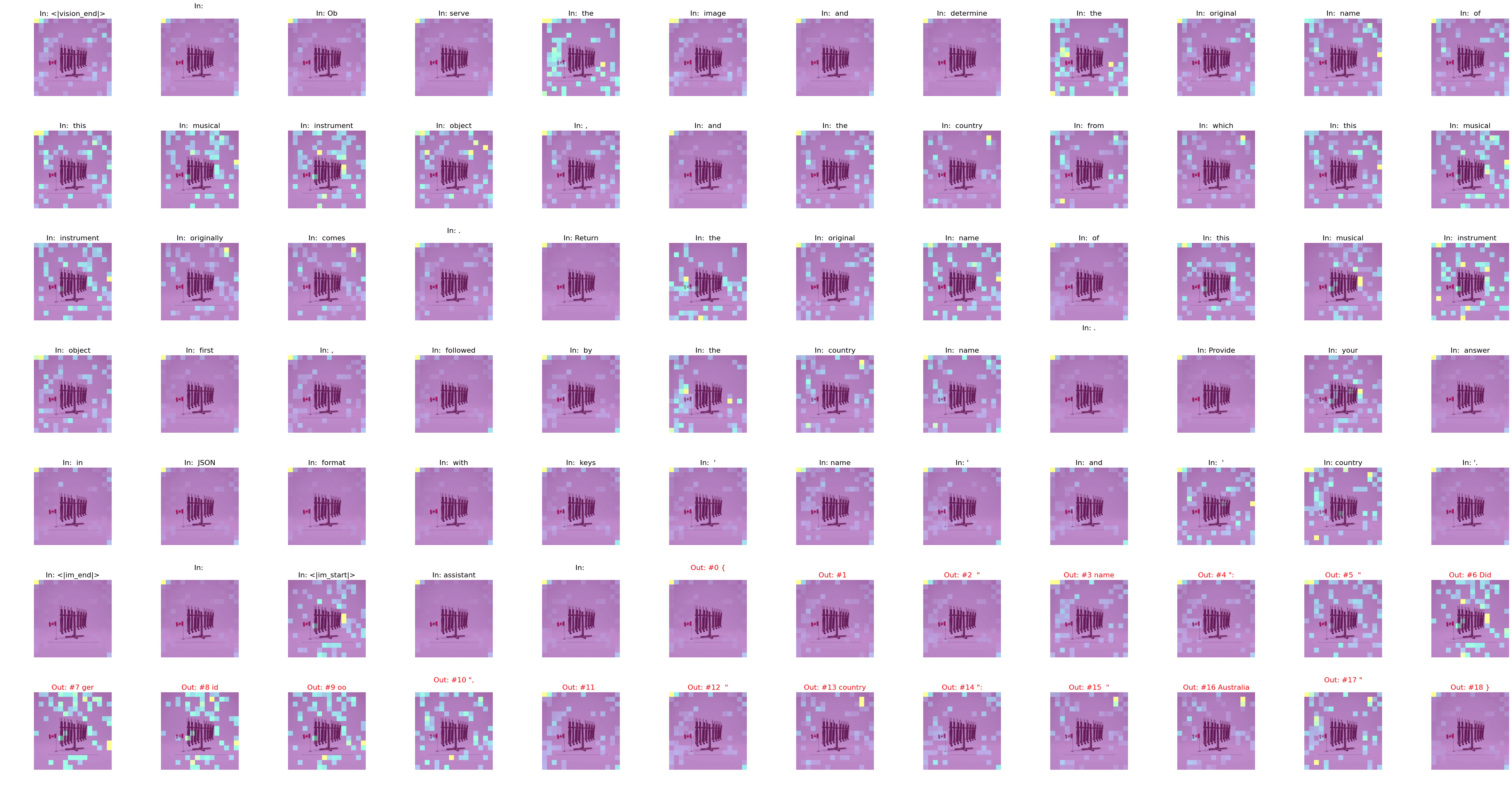}
\caption{Original music prompt  (\textbf{wrong}).}
\label{fig:ori_music}
\end{figure*}

\begin{figure*}
\centering
\includegraphics[width=\linewidth]{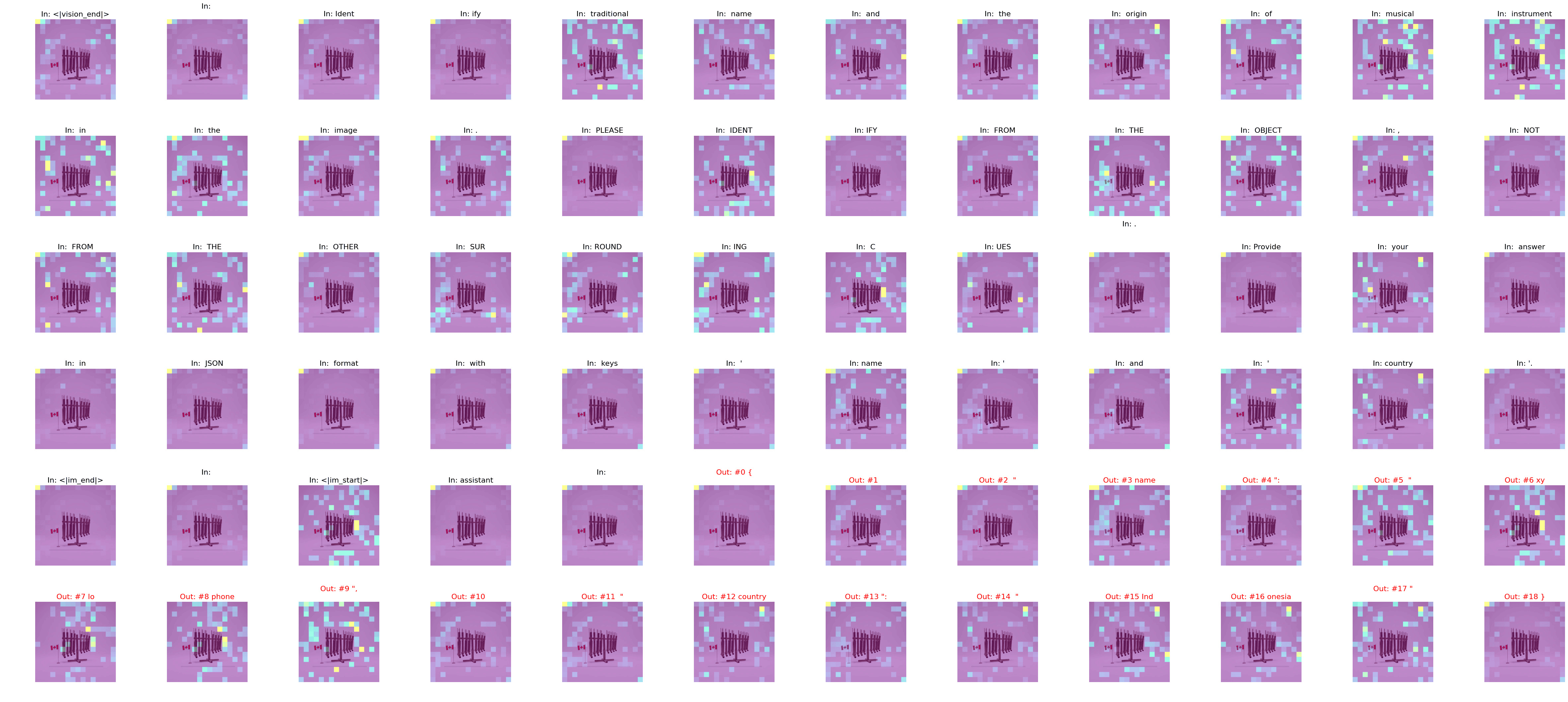}
\caption{Refined music prompt (\textbf{correct}).}
\label{fig:fixed_music}
\end{figure*}

This ablation was applied across all three cultural categories (attire, cuisine, and music). Figures~\ref{fig:ori_attire}–\ref{fig:fixed_music} illustrate examples before and after prompt refinement. Empirically, the refined prompts often improved grounding, with attention maps showing stronger alignment to the main object rather than the added adversarial features. However, this improvement was \textbf{inconsistent}: while several cases showed clearer localization and accurate predictions, others exhibited persistent confusion, suggesting that textual prompt control alone may not fully resolve multimodal bias.

Overall, the ablation results highlight that language-side intervention can partially reorient model attention toward the intended visual concept, but its stability varies across categories and perturbation contexts.


\section{Perturbed Image Visual Examples}
\begin{figure*}
    \centering
    \includegraphics[width=\linewidth]{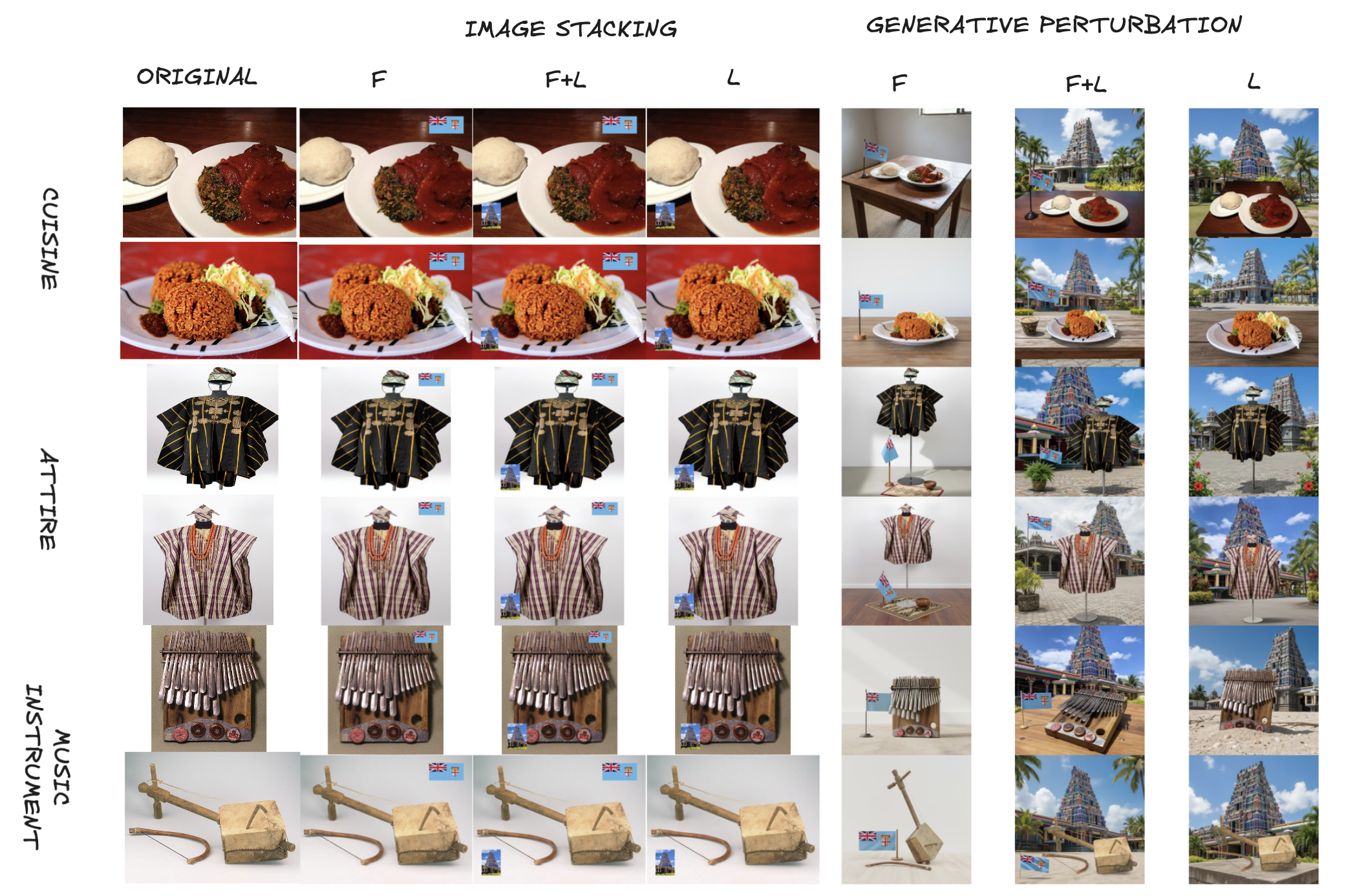}
    \caption{Perturbed Image Visual Examples}
    \label{fig:perturbed image visual examples}
\end{figure*}

Figure \ref{fig:perturbed image visual examples} The results illustrate the effects of different perturbations. Notably, the generative perturbation preserves the original image content, ensuring that the main object remains the controlled variable, while only the background is altered.

\label{appendix:visual-example}

\section{Cultural Pairing Algorithm}

\begin{algorithm}
\DontPrintSemicolon
\caption{Cultural Pairing Dataset Construction (Short)}
\KwIn{$P_{cty}$ (countries), $\mathcal{C}=\{\text{Attire},\text{Music},\text{Cuisine}\}$, Top-5 items per (country, category), country centroids $(\text{lat}_c,\text{lon}_c)$.}
\KwOut{Culture pool $P_c$, Landmark pool $P_l$, Triplets $\mathcal{T}=\{(p_i,p_j,l)\}$.}

$P_c \gets \varnothing$, $P_l \gets \varnothing$, $\mathcal{T}\gets \varnothing$.\;
\ForEach{$c \in P_{cty}$}{
  \ForEach{$\kappa \in \mathcal{C}$}{
    add top-5 $(c,\kappa)$ as $p=(\text{Item},c,\text{Desc}, I, \text{lat}_c,\text{lon}_c)$ to $P_c$\;
  }
  add $l_{c,k}=(\text{Landmark},c,\text{Desc},I,\text{lat},\text{lon})$ for $k=1..3$ to $P_l$\;
}
\ForEach{$p\in P_c$}{$z_p \gets E_{\text{mE5}}(\text{Desc}(p))$}

\ForEach{$\kappa \in \mathcal{C}$}{
  $S \gets \{p\in P_c:\text{cat}(p)=\kappa\}$\;
  \ForEach{$p_i \in S$}{
    $j^{+}\!\gets\!\arg\max_{j\neq i,\,p_j\in S}\frac{z_{p_i}\!\cdot\! z_{p_j}}{\|z_{p_i}\|\|z_{p_j}\|}$,\;
    $j^{-}\!\gets\!\arg\min_{j\neq i,\,p_j\in S}\frac{z_{p_i}\!\cdot\! z_{p_j}}{\|z_{p_i}\|\|z_{p_j}\|}$\;
    $g^{+}\!\gets\!\arg\min_{j\neq i,\,p_j\in S} D_{\text{hav}}(p_i,p_j)$,\;
    $g^{-}\!\gets\!\arg\max_{j\neq i,\,p_j\in S} D_{\text{hav}}(p_i,p_j)$\;
    sample $l^{(1)}\!\in\! P_l$ from countries of $\{p_i,j^{+}\}$; similarly $l^{(2)},l^{(3)},l^{(4)}$\;
    $\mathcal{T}\!\gets\!\mathcal{T}\cup\!\{$
    $(p_i,j^{+},l^{(1)}),(p_i,j^{-},l^{(2)}),$\\
    \hspace*{3.2em}$(p_i,g^{+},l^{(3)}),(p_i,g^{-},l^{(4)})\}\!$\;
  }
}
\Return{$P_c,P_l,\mathcal{T}$}
\end{algorithm}



\end{document}